\documentclass[10pt,twocolumn,letterpaper]{article}

\usepackage[pagenumbers]{cvpr} 

\definecolor{cvprblue}{rgb}{0.21,0.49,0.74}
\definecolor{rocket1}{HTML}{E1BAA0}
\definecolor{rocket2}{HTML}{D2866B}
\definecolor{rocket3}{HTML}{BC5157}
\definecolor{rocket4}{HTML}{8E335B}
\definecolor{rocket5}{HTML}{5F2D51}
\definecolor{rocket6}{HTML}{5F2D51}
\definecolor{rocket7}{HTML}{301E37}

\usepackage{array}
\newcolumntype{L}[1]{>{\raggedright\arraybackslash}m{#1}}
\newcolumntype{C}[1]{>{\centering\arraybackslash}m{#1}}

\usepackage[pagebackref,breaklinks,colorlinks,allcolors=cvprblue]{hyperref}

\usepackage{url}            
\usepackage{booktabs}       
\usepackage{amsfonts}       
\usepackage{nicefrac}       
\usepackage{microtype}      
\usepackage{xcolor}         
\usepackage{amsmath}
\usepackage{amsthm}
\usepackage{amssymb}
\usepackage{colortbl}
\usepackage{wrapfig}
\usepackage{graphicx}
\usepackage{enumitem}
\usepackage{mathtools}
\usepackage{tikz}
\usetikzlibrary{positioning}
\usepackage{makecell}
\usepackage{multirow}
\usepackage{minitoc} %

\usepackage{subcaption}
\usepackage{caption} 
\usepackage{pgffor}
\usepackage{etoolbox}
\usepackage{pgfplots}
\usepackage{arydshln}
\usepackage{amsfonts} 

\usepackage{tcolorbox}
\tcbuselibrary{skins, breakable, theorems}

\tcbset{
    thmbase/.style={
        enhanced jigsaw,
        breakable,
        colback=#1!10,
        coltitle=#1!70!black,
        fonttitle=\bfseries,
        description font=\normalfont\color{#1!70!black},
        separator sign={ },
        detach title,
        before upper={{\bfseries\color{#1!70!black}\tcbtitle .}\space}, 
        sharp corners,
        boxrule=0pt,
        leftrule=2pt,
        colframe=#1,
        left=6pt, right=6pt,
        top=2pt, bottom=2pt,
        boxsep=2pt,
        toptitle=0pt, bottomtitle=0pt,
    }
}

\newtcbtheorem[number within=section]{definition}{Definition}{thmbase=rocket2}{def}
\newtcbtheorem[]{theorem}{Theorem}{thmbase=rocket3}{thm}
\newtcbtheorem[]{lemma}{Lemma}{thmbase=rocket3}{lem}
\newtcbtheorem[]{prop}{Proposition}{thmbase=rocket3}{prop}
\newtcbtheorem[]{corollary}{Corollary}{thmbase=rocket3}{cor}
\newtcbtheorem[]{ex}{Example}{thmbase=rocket2, no counter}{ex}
\newtcbtheorem[]{remark}{Remark}{thmbase=rocket1, no counter}{rem}
\newtcbtheorem[]{note}{Note}{thmbase=rocket1}{note}
\newtcbtheorem[]{takeaway}{Takeaway}{thmbase=Goldenrod, no counter}{takeaway}

\renewenvironment{proof}[1][\proofname]{%
    \begin{tcolorbox}[thmbase=rocket3, colback=rocket3!4, title=#1, before skip=0pt]%
}{%
    \hfill\qedsymbol\end{tcolorbox}%
}

\def\confName{CVPR}
\def\confYear{2027}

\title{On the notion of missingness for path attribution explainability methods in medical settings: Guiding the selection of medically meaningful baselines}

\author{
Alexander Geiger\(^{1}\)
  \quad
  Lars Wagner\(^{1}\)
  \quad
  Daniel Rueckert\(^{2, 3}\)
  \quad
  Dirk Wilhelm\(^{1, 4}\)
  \quad
  Alissa Jell\(^{1, 4}\)
  \\[0.2em]
{\small
    \(^1\)Research Group MITI, Technical University of Munich} \quad
  {\small 
    \(^2\)Chair for AI in Healthcare and Medicine, Technical University of Munich} \\[-0.1em]
  {\small 
    \(^3\)Department of Computing, Imperial College London} \quad
  {\small 
    \(^4\)Department of Surgery, Technical University of Munich} \\
  {
    \tt\small \href{mailto:alexander.geiger@tum.de}{alexander.geiger@tum.de}
  }
}

\begin{document}
\twocolumn[{%
\renewcommand\twocolumn[1][]{#1}%
\maketitle 
\begin{center}
    \vspace{-0.25cm} 
    \centering
    \captionsetup{type=figure}
    \includegraphics[width=\linewidth]{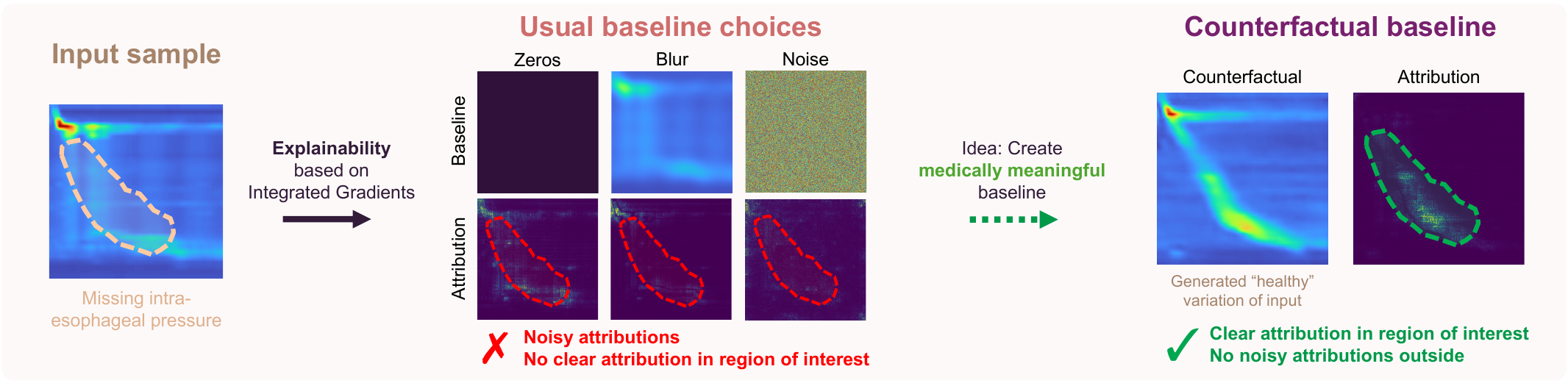}
    \vspace{-6mm}
  \caption{Effect of baseline choice on attribution quality for an pathological manometry input. Standard baselines (zeros, blurred input, random noise) yield spurious attributions outside the true region of interest, while our generated counterfactual baseline produces cleaner and more accurately localized attributions.
  }
    \label{fig:abstract}
\end{center}
}]
\begin{abstract}
The explainability of deep learning models remains a significant challenge, particularly in the medical domain where interpretable outputs are essential for clinical trust and transparency. Path attribution methods such as Integrated Gradients rely on a baseline that represents the absence of informative features, a notion commonly referred to as \emph{missingness}. However, commonly used baselines (such as all-zero inputs) are often semantically meaningless in medical contexts, where intensity values carry clinical significance. In this work, we revisit the notion of missingness for medical imaging, expose the limitations of standard baselines in this setting, and formalize a stricter missingness we term \emph{semantic missingness}: a baseline must not merely lack signal, but must represent a clinically plausible state in which the disease-related features are absent. This formulation motivates a counterfactual-guided approach to baseline selection, in which a synthetically generated counterfactual (i.e. a clinically normal variant of the pathological input) serves as a principled and semantically meaningful reference. We derive theoretical guarantees showing that counterfactual baselines yield more faithful attributions than standard alternatives, and empirically validate this with two complementary counterfactual generative models, a Variational Autoencoder and a diffusion model, though the concept is model-agnostic and compatible with any suitable counterfactual method. Our experiments show that counterfactual baselines produce consistently more faithful and medically relevant attributions across three diverse medical datasets, outperforming both standard baseline choices and other related methods. Lastly, we also compare against using the counterfactual directly as an explanation (an established paradigm in its own) and show that employing it as a baseline for Integrated Gradients yields superior results, thereby effectively bridging two separate explainability paradigms.
\end{abstract}    
\section{Introduction}

The integration of Artificial Intelligence (AI) into medical practice is rapidly transforming the landscape of healthcare, offering powerful tools for diagnostics, prognosis, and treatment planning. However, despite their accuracy, these models often operate as ``black boxes'', offering little insight into how specific predictions are made. To this end, the ability to interpret and understand the internal decision-making processes of AI models is of great importance in the medical domain and many different approaches have been developed over the years \cite{hossainExplainableAIMedical2025a, brandenburgCanSurgeonsTrust2025, sunExplainableArtificialIntelligence2025}.
Among the most widely used explainability methods is Integrated Gradients (IG)~\cite{sundararajanAxiomaticAttributionDeep2017}, a path attribution technique that satisfies several desirable axioms~\cite{sundararajanAxiomaticAttributionDeep2017, lundstromRigorousStudyIntegrated2022}. A critical component of IG, as well as other explainability concepts like DeepLift~\cite{shrikumarLearningImportantFeatures2017} and GradientSHAP~\cite{lundbergUnifiedApproachInterpreting2017}, is the choice of a baseline reference, which ideally represents an absence (``missingness'') of any relevant features. This concept of missingness is central to interpretability research~\cite{sturmfelsVisualizingImpactFeature2020}.
A common baseline is a zero-valued input, which corresponds to a black image in computer vision tasks. However, since IG is defined to attribute differences in model output between the baseline and the input without regard to the semantic appropriateness of the baseline, such choices can lead to misleading attributions (e.g., IG is `blind' to the baseline color). Thus, alternative baselines such as blurred inputs or random noise have been proposed~\cite{fongInterpretableExplanationsBlack2017}.
Yet, these alternatives often lie outside the data distribution the model was trained on, potentially resulting in erroneous attributions due to unreliable gradients in regions of input space the model has never encountered. Therefore, \citet{erionImprovingPerformanceDeep2021} introduced Expected Gradients (EG), an extension of IG that samples multiple baselines from the training distribution and averages the resulting attributions.

Despite this advancement, we show in this work that even these data-aware baselines produce poor attributions in certain medical contexts. For example, when the class-defining feature is itself a form of missingness, standard baselines fail to capture the intended contrast. This can be observed, for instance, in esophageal manometry recordings, where pathology can be characterized by the absence of normal pressure patterns, or in chest X-rays of pneumothorax patients, where the condition can manifest as a missing lung boundary or contrast. These shortcomings indicate that a stricter notion of missingness for the baseline is required: one that does not merely encode an arbitrary or signal-free input, but also represents a clinically plausible state in which the disease-related features are genuinely absent. This motivates the central idea of our work, replacing conventional baselines with synthetically generated \emph{counterfactual baselines} that reflect a clinically normal variant of the pathological input, thereby providing a principled and semantically meaningful reference for attribution. Fig.~\ref{fig:abstract} illustrates how this aligned notion of missingness yields cleaner and more clinically relevant attributions than conventional baselines.

To our knowledge, this is the first work to systematically examine missingness in the medical context and its implications for baseline selection. We summarize our contributions as follows:
\begin{itemize}[topsep=2pt, itemsep=3pt, parsep=0pt, leftmargin=*]
    \item \emph{Formalization of semantic missingness.} We revisit the notion of missingness in the context of medical imaging, expose the limitations of standard baselines, and introduce \emph{semantic missingness} as a stricter requirement: a baseline must not merely lack signal, but represent a clinically plausible state in which disease-related features are absent, thus motivating counterfactual baselines.
    \item \emph{Theoretical analysis.} We derive formal guarantees showing why counterfactual baselines yield more faithful attributions, providing a principled justification for our concept.
    \item \emph{Empirical evaluation across three medical datasets.} We compare a broad range of baselines and related concepts against our counterfactual baselines on three diverse medical imaging datasets, and demonstrate consistent improvements in both faithfulness and clinical relevance.
    \item \emph{Bridging counterfactual explanations and path attribution.} We show that using a generated counterfactual as the baseline for Integrated Gradients outperforms using the counterfactual directly as the explanation itself (an established paradigm in its own), thereby connecting two previously distinct branches of the explainability literature.
\end{itemize}

\paragraph{Related literature}
Explainability methods have gained significant attention in recent years, particularly as AI systems are increasingly deployed in high-stakes domains. Broad overviews of explainable artificial intelligence (XAI) are provided by \citet{minhExplainableArtificialIntelligence2022} and \citet{angelovExplainableArtificialIntelligence2021}, who review general techniques and motivations for interpretability across domains. In medical settings, the need for transparency and clinical trust makes interpretability especially critical \cite{brandenburgCanSurgeonsTrust2025}, and several surveys discuss both the domain-specific challenges and the adaptation of general techniques \cite{sunExplainableArtificialIntelligence2025, hossainExplainableAIMedical2025a, vanderveldenExplainableArtificialIntelligence2022, chaddadSurveyExplainableAI2023}.

Generative models have recently emerged as powerful tools for improving the interpretability of medical AI models, for example by translating images between domains \cite{baumgartnerVisualFeatureAttribution2018} or by encoding input features into structured latent representations \cite{bassICAMInterpretableClassification2020} that can be traversed to observe and explain corresponding model outputs \cite{cloughGlobalLocalInterpretability2019, difolcoInterpretableRepresentationLearning2024}. Closely related is the use of counterfactual explanations, which provide generated contrastive inputs that lead to different model outcomes and thereby offer an intuitive way to visualize the factors influencing a model's decision \cite{schutteUsingStyleGANVisual2021, mertesGANterfactualCounterfactualExplanations2022, pegiosCounterfactualExplanationsRiemannian2024, atadCounterfactualExplanationsMedical2024, tanyelKnownRealityExploiting2025, nageshExplainingMachineLearning2023, wengFastDiffusionBasedCounterfactuals2025, mothilalExplainingMachineLearning2020, guyomardVCNetSelfexplainingModel2022}. While counterfactuals provide such intuitive explanations on their own, we show in this work that they can be leveraged even more effectively when used as baselines for path attribution methods such as IG, which remain a cornerstone of explainability for understanding feature-level contributions.

A complementary line of research addresses noisy IG attributions by adapting the integration path. \citet{kapishnikovGuidedIntegratedGradients2021} introduce Guided Integrated Gradients, which optimizes the path along the model's output surface, while \citet{zhuoIG2IntegratedGradient2024a} extend this by jointly deriving the path and baseline from a counter-class reference. Further geometric approaches have also been explored in this regard \citep{zaherManifoldIntegratedGradients2024}. While these methods improve upon the linear path of standard IG, they treat the baseline as fixed and inherited, leaving semantically meaningful baseline selection unaddressed.
Here, a separate line of work examines the choice of baseline itself. \citet{dravidMedXGANVisualExplanations2022} propose Latent Integrated Gradients (LIG), which uses a GAN-generated counterfactual as the baseline and computes IG in the generator's latent space. However, as LIG operates entirely in the generated domain, the actual input is never directly explained, limiting clinical interpretability. Most closely related to our work, yet in the tabular data domain, \citet{duellCounterfactualIntegratedGradientsCounterfactual2023} combine IG with counterfactual baselines for tabular medical records. However, they treat the counterfactual as a given explanation rather than as a principled baseline choice, provide no comparison to alternative baselines, and no formal attribution quality assessment, thus leaving entirely open whether and why counterfactuals constitute a better baseline than standard alternatives.

Our work addresses these gaps in several ways. We provide a formal framework for why counterfactual baselines are particularly suited to medical settings, systematically compare against a broad range of alternative baselines and related approaches across three medical datasets, as well against using counterfactuals alone as explanations. Crucially, we build directly on the standard IG formulation rather than introducing adaptive paths or modified attribution computations, thereby staying faithful to the original IG definition and automatically inheriting all of its several attribution axioms.

\section{Motivation}
IG is an attribution method that assigns importance to each input feature by integrating the gradients of the model’s output with respect to the inputs along a path from a baseline input (ideally representing an absence of any features of the target class) to the actual input. The IG for input $x \in \mathbb{R}^n$ and baseline $\hat{x} \in \mathbb{R}^n$ is defined as:
\(
\mathrm{IG}_i(x, \hat{x}) = (x_i - \hat{x}_i) \cdot \int_{\alpha=0}^1 \frac{\partial \mathcal{C}\left(\hat{x} + \alpha \cdot (x - \hat{x} )\right)}{\partial x_i} \, d\alpha,
\)
where $\mathcal{C}: \mathbb{R}^n \rightarrow \mathbb{R}$ is the classifier function (e.g., the output logit or probability for a specific class) and $\alpha$ is a scalar interpolating from $\hat{x}$ to $x$.

\paragraph{Missingness and Baseline Semantics}
IG explains a prediction relative to a baseline $\hat{x}$ via
\(
\sum_{i=1}^n \mathrm{IG}_i(x;b)
=
\mathcal{C}_y(x) - \mathcal{C}_y(b),
\)
where the baseline represents the state in which the target concept $y$ is ``absent'', defining basic missingness.

\begin{definition}{(Basic missingness)}{}
A baseline $\hat{x}$ satisfies \emph{basic missingness} for class $y$ if \(\mathcal{C}_y(\hat{x}) \approx 0\), i.e. the output logit is near neutral for class $y$.
\end{definition}
In this sense, basic missingness just means that the baseline does not contain discriminative information for class $y$ under the model. Standard baselines are designed with this notion in mind: zero, blurred, and noise inputs all aim to produce low classifier confidence for any specific class, and EG approximates the same goal by averaging over training-data baselines. In medical imaging, however, low classifier evidence alone does not guarantee a meaningful absence of the concept. We accordingly introduce a stricter notion that demands not only the lack of class evidence, but also semantic plausibility and structural relation to the input.

\begin{definition}{(Semantic missingness)}{}
A baseline $\hat{x}$ satisfies \emph{semantic missingness} for class $y$ if (1) \(\mathcal{C}_y(\hat{x}) \approx 0\), (2) \(\hat{x} \in \mathcal{M} \), and (3) \(\|\hat{x} - x\|_2\) is small.
\end{definition}
Intuitively, a baseline $\hat{x}$ that satisfies semantic missingness represents a semantically plausible input that closely resembles $x$ but no longer contains evidence for class $y$, i.e. a counterfactual.

\paragraph{Baseline choice in medical imaging: limitations and the case for counterfactuals} 
With this definition in mind, several distinct shortcomings of standard baselines and corresponding advantages of counterfactual baselines emerge.

\textit{Semantically meaningful vs.\ merely null reference.} Standard baselines aim only for low classifier evidence (condition 1), but the resulting input is typically not a semantically valid state: a zero or noise image is not something the model has ever encountered as a plausible medical input. Counterfactual baselines, by construction, represent a clinically normal variant of $x$ and therefore provide a reference state that is meaningful in the diagnostic sense, not merely null with respect to the classifier.

\textit{On- vs.\ off-manifold reference and interpolation paths.} Black, blurred, and noisy baselines lie far from the data manifold (violating condition 2), and EG's training-data baselines, although individually on-manifold, are anatomically unrelated to the input and therefore induce paths that pass through implausible intermediate states. Such paths are particularly problematic when pathology manifests as a subtle perturbation of normal tissue, and the effect of out-of-distribution intermediate inputs on attribution quality is poorly understood \cite{sturmfelsVisualizingImpactFeature2020, hookerBenchmarkInterpretabilityMethods2019, fongInterpretableExplanationsBlack2017}. 
A counterfactual $x_{\text{normal}}$ satisfies the manifold condition by construction and, because it is conditioned on $x$ (condition 3), induces an interpolation path $\gamma(\alpha) = x_{\text{normal}} + \alpha(x - x_{\text{normal}})$ that transitions smoothly from a clinically normal to a pathological state, more likely to remain on the manifold.

\textit{Accidental matching of pathological values.} Standard baselines are agnostic to how pathology manifests in a given input and may coincidentally match pathological intensity values. For instance, low intensities encode absent peristalsis in manometry. The result is not merely a lack of neutrality but active suppression of attribution signal in the regions of interest, since the contrast term $(x_i - \hat{x}_i)$ vanishes precisely where it should be largest. Counterfactual baselines are explicitly constructed to differ from the input in pathological regions, eliminating this failure mode and concentrating gradient signal where it is diagnostically meaningful.

\section{Counterfactual baselines}
We first formalize the notion of a perfect counterfactual baseline in the context of path-based attribution methods and show that, under this idealization, Integrated Gradients assigns non-zero attribution only to pathological coordinates, i.e. achieves a perfect attribution map with no spurious attributions.
Let \(\mathcal{C} : \mathbb{R}^n \to \mathbb{R}\) be a differentiable function representing the scalar prediction of interest (e.g., the logit for a disease class). Let $x \in \mathbb{R}^n$ be a fixed input and let \(S \subseteq \{1,\dots,n\}\) denote the (unknown) set of \emph{pathological coordinates}, i.e., the coordinates where the input differs from a corresponding healthy, pathology-free version.

\begin{definition}{(Perfect Counterfactual Baseline)}{}
A baseline $\text{CF}^{*} \in \mathbb{R}^n$ is called a \emph{perfect counterfactual baseline} if $\mathcal{C}(\text{CF}^{*})$ corresponds to a neutral prediction, there exists a minimal causal index set $S \subseteq \{1,\dots,n\}$ such that \(x_i = \text{CF}^{*}_i \quad \text{for all } i \notin S;\), and the prediction difference \(\mathcal{C}(x) - \mathcal{C}(\text{CF}^{*})\) depends only on coordinates in $S$.
\end{definition}

Intuitively, $\text{CF}^{*}$ differs from $x$ only in those features causally responsible for the prediction. In this case, we can show that all attribution mass is concentrated on the causal coordinates $S$ only.

\begin{prop}{(Perfect Attribution under Perfect Counterfactual Baseline)}{perfect_attribution}
Let $\mathcal{C}$ be continuously differentiable and let $\text{CF}^{*}$ be a perfect counterfactual baseline for $x$. 
Then (1) $\mathrm{IG}_i(x;\text{CF}^{*}) = 0$ for all $i \notin S$ and (2) $\sum_{i \in S} \mathrm{IG}_i(x;\text{CF}^{*}) = \mathcal{C}(x) - \mathcal{C}(\text{CF}^{*})$.
\end{prop}
This follows directly from the definitions of IG and the Perfect Counterfactual Baseline (see appendix~\ref{sec:appendix_prop1}).
However, in practice, a perfect counterfactual is likely not available or hard to compute. We therefore also consider the more realistic case of approximate counterfactual baselines.

\paragraph{Approximate counterfactuals}

We show that, under mild assumptions, if an approximate counterfactual baseline \(\tilde x\) is closer to the perfect counterfactual \(\text{CF}^{*}\) than another (standard) baseline \(b\) is, then the corresponding attribution \(\mathrm{IG}(x; \tilde x)\) is closer to the ideal attribution \(\mathrm{IG}(x; \text{CF}^{*})\) in terms of error upper bound than \(\mathrm{IG}(x; b)\) is.

\begin{prop}{(Closeness to True Counterfactual Implies Lower Attribution Error Upper Bound)}{approx-counterfactual-better}
Let $b \in \mathbb{R}^d$ be an arbitrary baseline. Define
\(\delta_b := \| b - \text{CF}^{*} \|_2\) and \(C := \frac{L}{2}\|x - \text{CF}^{*}\|_2 + G_\infty\). Under assumptions (A1) and (A2) (see appendix \ref{sec:appendix_prop2}), for any baseline $b$,
\[
\|\mathrm{IG}(x;b) - \mathrm{IG}(x;\text{CF}^{*})\|_2
\le
C \, \delta_b.
\]
Consequently, for two baselines $b_1, b_2$, if
\(\|b_1 - \text{CF}^{*}\|_2 < \|b_2 - \text{CF}^{*}\|_2,\)
then their corresponding worst-case upper bounds satisfy
\(C \|b_1 - \text{CF}^{*}\|_2
<
C \|b_2 - \text{CF}^{*}\|_2\).
Hence, a baseline closer to the true counterfactual admits a strictly tighter theoretical upper bound on the attribution deviation.
\end{prop}

\begin{figure}
    \centering
    \includegraphics[width=0.8\linewidth]{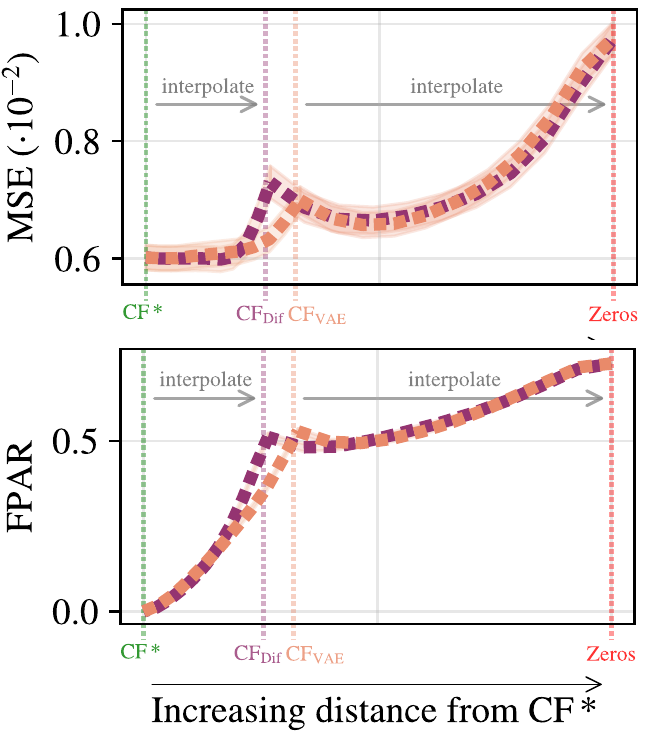}
    \caption{
    MSE and FPAR when increasing the distance to the perfect counterfactual $\mathrm{CF^*}$ by interpolating to the approximated counterfactual baselines and then to the zeros baseline.
    }
    \label{fig:distance_plot}
\end{figure}

Proposition~\ref{prop:approx-counterfactual-better} formalizes the intuition that Integrated Gradients is stable under small perturbations of the baseline (see appendix~\ref{sec:appendix_prop2} for the full derivation).
Thus, if we regard the perfect counterfactual baseline \(\text{CF}^{*}\) as defining an ideal, pathology-only attribution (Proposition~\ref{prop:perfect_attribution}), then any approximate counterfactual baseline $\hat{\text{CF}}$ that is close to \(\text{CF}^{*}\) in input space will, by design, yield attributions that are close to this ideal.
Conversely, standard baselines \(b\) that are far from \(\text{CF}^{*}\), and typically differ from \(x\) in large healthy regions, have a larger upper bound, which allows attributions to potentially deviate substantially from the ideal attribution, often exhibiting spurious importance outside the pathological region. 

In order to demonstrate this theoretical result empirically in our setting, we constructed a synthetic dataset in which the true counterfactual $\text{CF}^{*}$ is known by design, enabling direct measurement of attribution quality as a function of baseline distance. As shown in Fig.~\ref{fig:distance_plot}, attribution quality degrades gradually as the baseline moves from $\text{CF}^{*}$ through our generated approximations $\text{CF}_{\text{Dif.}}$ and $\text{CF}_{\text{VAE}}$ toward the zeros baseline, substantiating the theoretical proximity-attribution relationship. We refer to Appendix~\ref{sec:appendix_proximity} for full details of the synthetic experiment.
\subsection{Method: Generation of Counterfactual Baseline}

We propose using generated counterfactual images as baselines for path attribution methods. While we instantiate this concept with two complementary generative approaches (a VAE-based and a diffusion-based method), the framework is model-agnostic: any counterfactual generation technique that produces anatomically plausible, disease-free variants of a pathological input can be substituted, with the potential for further gains using more advanced methods. We demonstrate our concept using two generative approaches that differ in their inductive biases and guidance mechanisms, providing evidence that the benefits of counterfactual baselines are not tied to a specific generation paradigm.

The \textbf{VAE-based approach} encodes the input $x$ into a latent representation and optimizes the latent code toward the healthy class using classifier feedback, before decoding to obtain the counterfactual $\hat{\text{CF}}$. This constitutes a classifier-guided generation strategy operating in a compressed latent space.

The \textbf{diffusion-based approach} instead adds noise to the input up to a fixed timestep and then denoises it using a classifier-free guided diffusion model conditioned on the healthy class, steering the reconstruction away from pathological features without requiring an explicit latent optimization loop.
Full implementation details for both approaches are provided in Appendix~\ref{sec:appendix_counterfactual_generation}. Together, they cover both classifier-guided and classifier-free generation strategies, further supporting the method-agnostic nature of our framework. The classifier-independent approach also rules out any potential circularity issues a classifier-guided approach might have. Additionally, while our framework does not require generated counterfactuals to be perfect (Proposition~\ref{prop:approx-counterfactual-better}), we nevertheless assess their quality to rule out unintended adversarial or off-manifold artifacts. Here we observe that the generated counterfactuals indeed substantially shift inputs toward the real healthy distribution while concentrating modifications in clinically relevant regions. We report the quantitative realism scores and their correlation to attribution quality metrics in Appendix~\ref{sec:cf_realism}.

\subsection{Attribution with Counterfactual Baselines}

Given a generated counterfactual $\hat{\text{CF}}$ for input $x$, the attribution is computed by using $\hat{\text{CF}}$ as the baseline in the standard IG formulation

\(
\mathrm{IG(CF)}_i(x) = (x_i - \hat{\text{CF}}_i) \cdot \int_{\alpha=0}^1 \frac{\partial \mathcal{C}\left(\hat{\text{CF}} + \alpha \cdot (x - \hat{\text{CF}})\right)}{\partial x_i} \, d\alpha
\)

Both generative approaches also admit a natural extension to Expected Gradients (EG)~\cite{erionImprovingPerformanceDeep2021} by constructing a distribution of diverse yet semantically consistent counterfactual baselines. For the VAE-based approach, this is achieved by sampling perturbations around the optimized latent code $z^*$, e.g.\ $z \sim \mathcal{N}(z^*(x),\, \sigma^2_\phi(x)\, I)$, and decoding each sample into a distinct counterfactual. For the diffusion-based approach, diversity is introduced by applying independent noise realizations at each denoising step using a stochastic DDIM sampler with $\eta = 0.3$, yielding a set of varied yet plausible healthy reconstructions. In both cases, the EG attribution is:
\begin{equation}
\begin{aligned}
\mathrm{EG(CF)}_i(x) = \mathbb{E}_{\hat{\text{CF}} \sim p_{\text{CF}}(x)} \biggl[ (x_i - \hat{\text{CF}}_i) \cdot \\\int_{\alpha=0}^1 \frac{\partial \mathcal{C}\left(\hat{\text{CF}} + \alpha \cdot (x - \hat{\text{CF}}_i)\right)}{\partial x_i} \, d\alpha \biggr]
\end{aligned}
\end{equation}
where $p_{\text{CF}}(x)$ denotes the distribution over counterfactual baselines induced by either the VAE latent perturbations or the diffusion noise realizations.

\begin{figure*}
\centering
\includegraphics[width=\textwidth]{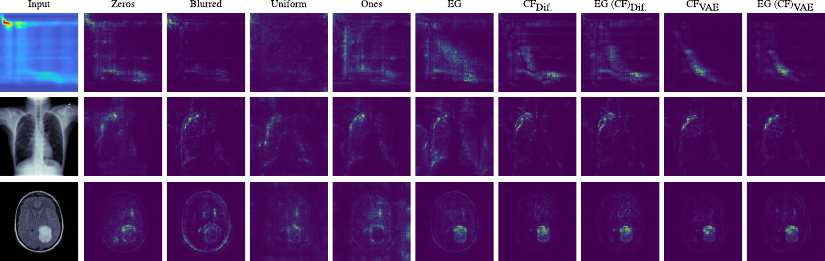}
\caption{
Examples of attributions obtained using different baselines.
}
\label{fig:examples}
\end{figure*}

\section{Evaluation}
\begin{table}
\vspace{-1.3em}
  \scriptsize
  \centering
  \caption{Description of compared baselines}
 \begin{tabular}{@{\hskip 0.4em} l @{\hskip 1.4em} l @{\hskip 0.4em}}
    \toprule
    \textbf{Baseline}  & \textbf{Description} \\
    \midrule
    \textsf{\scriptsize Zeros} & Tensor of zeros \\
   \textsf{\scriptsize Blurred} & Blurred input using gaussian filter ($\sigma = 20$)  \\
    \makecell[l]{\textsf{\scriptsize Uniform} \\ \strut} & \makecell[l]{Random noise input, where each pixel drawn \\ from Uniform distribution} \\
    \textsf{\scriptsize Ones} & Tensor of ones (as counterpart to \textsf{\scriptsize Zeros}) \\
    \makecell[l]{\textsf{\scriptsize EG} \\ \strut} & \makecell[l]{Expected Gradients using a set of 50 randomly \\ drawn training samples as baselines} \\
    \hdashline[2pt/2pt]
    \noalign{\vskip 0.3ex}
    \makecell[l]{\textsf{\scriptsize CF\textsubscript{Dif.}} \\ \strut} & \makecell[l]{The Diffusion-generated counterfactual \\ representing the normal case} \\
    \makecell[l]{\textsf{\scriptsize EG (CF)\textsubscript{Dif.}} \\ \strut} & \makecell[l]{Expected Gradients using a set of \\ 50 Diffusion-generated counterfactuals} \\
    \makecell[l]{\textsf{\scriptsize CF\textsubscript{VAE}} \\ \strut} & \makecell[l]{The VAE-generated counterfactual \\ representing the normal case} \\
    \makecell[l]{\textsf{\scriptsize EG (CF)\textsubscript{VAE}} \\ \strut} & \makecell[l]{Expected Gradients using a set of \\ 50 VAE-generated counterfactuals} \\
    \bottomrule
    \end{tabular}
  \label{tab:baselines_description}
  \vspace{-2em} 
\end{table} 

Evaluating explainability methods, especially attribution maps, is challenging, as it is often unclear whether observed errors stem from the model or the attribution technique itself, prompting the identification of key limitations and the development of various evaluation strategies in recent years \cite{dabkowskiRealTimeImage2017, anconaBetterUnderstandingGradientbased2018a, hookerBenchmarkInterpretabilityMethods2019, adebayoSanityChecksSaliency2018, yehInfidelitySensitivityExplanations2019, kindermansUnreliabilitySaliencyMethods2019, kimHIVEEvaluatingHuman2022, lageEvaluationHumanInterpretabilityExplanation2019, narayananHowHumansUnderstand2018, kadirEvaluationMetricsXAI2023}.

To evaluate our proposed counterfactual-based baseline approaches, we compare them against standard baseline choices on three diverse medical data sets. Using multiple metrics, we demonstrate that our method constantly produces the most informative attributions by better highlighting the most relevant pixels. A detailed overview of the compared baseline choices is provided in Table~\ref{tab:baselines_description}. All experiments were implemented in PyTorch and conducted on a single NVIDIA L40S GPU.

\paragraph{Data sets}

We evaluate the baselines on three medical data sets: manometry, chest X-ray, brain MRI. The data sets contain class labels and segmentation masks for pathological region localization. Images are resized to $512 \times 512$ and normalized to $[0, 1]$. Evaluation is based on pathological test samples with available segmentation masks. Dataset details, splits, and sample sizes are described in appendix \ref{sec:appendix_dataset}. For a qualitative impression, Fig.~\ref{fig:examples} shows an example input from each data set with corresponding attribution maps for the various baselines we compare.

\subsection{Metrics}
We use multiple metrics to evaluate the performance of the different baselines. Before evaluating, the attributions are capped at the 99.9th percentile to avoid outlying high-intensity attributions to dominate the visualizations and evaluations. 

\textbf{Attribution intensity distribution}\label{sec_metrics_attribution_distribution}
To compare how cleanly and how medically plausibly the different baselines attribute relevance, we examine the distribution of attribution intensities separately inside and outside the ground-truth segmentation masks. A medically meaningful attribution map should place high attribution intensities predominantly inside the mask and only low intensities outside it. We use the \textsf{\small Zeros} baseline as the reference: for each baseline, we compute the intensity histogram and report, per bin, the $\log_2$ ratio of its frequency to that of the \textsf{\small Zeros} baseline. Positive values thus indicate that a baseline produces more attributions of that intensity, and negative values indicate fewer.

\textbf{Overlap of attributions with ground truth segmentation mask}
In general, attribution maps should align with regions of interest, making their overlap with ground truth segmentation masks an established measure of plausibility \cite{dabkowskiRealTimeImage2017, bylinskiiWhatDifferentEvaluation2019, kapishnikovGuidedIntegratedGradients2021, arrasCLEVRXAIBenchmarkDataset2022}. To measure the overlap, we treat the attribution maps as predictions and the binary segmentation masks as targets. We then compute both the Precision-Recall AUC (PR-AUC) as well as the Mean Squared Error (MSE) as two measures of alignment.
In addition, to assess attribution noise outside the relevant region, we report the false positive attribution rate (FPAR), defined as the fraction of total attribution mass that falls outside the ground-truth mask. A lower FPAR therefore indicates a more focused attribution map that assigns less relevance to non-relevant regions. FPAR is the complement of the established relevance mass accuracy metric proposed by \citet{arrasCLEVRXAIBenchmarkDataset2022} (i.e., FPAR = 1 - relevance mass accuracy).

\textbf{Ablation tests}\label{sec_metrics_ablation}
Ablation tests are another intuitive way to assess attribution quality: pixels are replaced in order of their attributed importance, and a faster drop in classifier confidence (i.e., a lower curve) indicates a better attribution map \cite{sturmfelsVisualizingImpactFeature2020, ghorbaniInterpretationNeuralNetworks2019}. However, the drop may stem from the imputations producing out-of-distribution inputs rather than from the removed pixels being important \cite{hookerBenchmarkInterpretabilityMethods2019}, and it is unclear which replacement values are truly neutral. Both issues are especially relevant in our setting, where the classifier detects subtle patterns in medical images that pixel replacements easily distort.
We therefore perform mass-center and top-$k$ ablation tests with multiple replacement values. As a reference, we construct a mask-restricted map by setting all \textsf{\small Zeros} attributions outside the ground-truth mask to zero, so that by construction it contains no false attributions. An ablation test that measures domain-specific faithfulness should yield low curves for this map, yet it does not on any data set. Ablation tests thus capture different properties: the classifier may draw on information outside the annotated region, or the perturbation-induced distribution shift may dominate the confidence drop. As our experiment cannot disentangle these effects, we do not treat ablation tests as a measure of semantically meaningful attribution faithfulness in our medical settings. Details and full results, including the mask-restricted map, are given in Appendix~\ref{sec:app_ablation}.

\setlength\dashlinedash{1pt}
\setlength\dashlinegap{1pt}
\setlength\arrayrulewidth{0.6pt}
\begin{table*}
  \footnotesize
  \centering
  \caption{Aggregated scores (mean $\pm$ se) measuring the overlap between attribution map and ground truth mask; p $<$ 0.01 (*); The \textbf{best} and three next best scores are highlighted.}
    \setlength{\tabcolsep}{1.65pt}

    \begin{tabular}{
  l
  >{\hskip 0.02em}c<{\hskip 0.02em}
  >{\hskip 0.02em}c<{\hskip 0.02em}
  >{\hskip 0.02em}c<{\hskip 0.02em}
  >{\hskip 0.02em}c<{\hskip 0.02em}
  >{\hskip 0.02em}c<{\hskip 0.02em}
  >{\hskip 0.02em}c<{\hskip 0.02em}
  >{\hskip 0.02em}c<{\hskip 0.02em}
  >{\hskip 0.02em}c<{\hskip 0.02em}
  >{\hskip 0.02em}c<{\hskip 0.02em}
}
    \toprule
     & \multicolumn{3}{c}{\textbf{Manometry}} & \multicolumn{3}{c}{\textbf{Chest X-ray}} & \multicolumn{3}{c}{\textbf{Brain MRI}} \\
    \cmidrule(lr){2-4} \cmidrule(lr){5-7} \cmidrule(lr){8-10}
    \textbf{Baseline} & \multicolumn{1}{c}{MSE $\downarrow$ {\fontsize{4}{5}\selectfont{$\left(\cdot10^{\text{-}2}\right)$}}} & \multicolumn{1}{c}{PR-AUC $\uparrow$} & \multicolumn{1}{c}{FPAR $\downarrow$} & \multicolumn{1}{c}{MSE $\downarrow$ {\fontsize{4}{5}\selectfont{$\left(\cdot10^{\text{-}2}\right)$}}} & \multicolumn{1}{c}{PR-AUC $\uparrow$} & \multicolumn{1}{c}{FPAR $\downarrow$} & \multicolumn{1}{c}{MSE $\downarrow$ {\fontsize{4}{5}\selectfont{$\left(\cdot10^{\text{-}2}\right)$}}} & \multicolumn{1}{c}{PR-AUC $\uparrow$} & \multicolumn{1}{c}{FPAR $\downarrow$} \\
    \midrule
    \textsf{\footnotesize Zeros}
  & 4.50 $\pm$ 0.29 & 0.19 $\pm$ 0.02 & 0.85 $\pm$ 0.01
  & 1.52 $\pm$ 0.07 & 0.12 $\pm$ 0.01 & 0.92 $\pm$ 0.00
  & 2.51 $\pm$ 0.12 & 0.39 $\pm$ 0.01 & \cellcolor{rocket2!15} 0.65 $\pm$ 0.01 \\

\textsf{\footnotesize Blurred}
  & 4.80 $\pm$ 0.30 & 0.12 $\pm$ 0.01 & 0.89 $\pm$ 0.01
  & \cellcolor{rocket2!45}1.33 $\pm$ 0.08 & \cellcolor{rocket2!45}0.14 $\pm$ 0.01 & \cellcolor{rocket2!45}0.87 $\pm$ 0.01
  & 2.90 $\pm$ 0.12 & 0.21 $\pm$ 0.01 & 0.81 $\pm$ 0.01 \\

\textsf{\footnotesize Uniform}
  & 4.85 $\pm$ 0.29 & 0.09 $\pm$ 0.01 & 0.92 $\pm$ 0.01
  & 1.59 $\pm$ 0.07 & 0.07 $\pm$ 0.00 & 0.94 $\pm$ 0.00
  & 2.80 $\pm$ 0.12 & 0.21 $\pm$ 0.00 & 0.82 $\pm$ 0.00 \\

\textsf{\footnotesize Ones}
  & 4.55 $\pm$ 0.29 & 0.15 $\pm$ 0.01 & 0.88 $\pm$ 0.01
  & \cellcolor{rocket2!15} 1.36 $\pm$ 0.07 & \cellcolor{rocket2!60}\textbf{0.15 $\pm$ 0.01} & \cellcolor{rocket2!15} 0.90 $\pm$ 0.01
  & 2.91 $\pm$ 0.12 & 0.21 $\pm$ 0.01 & 0.84 $\pm$ 0.00 \\

\textsf{\footnotesize EG}
  & 4.45 $\pm$ 0.25 & 0.22 $\pm$ 0.02 & \cellcolor{rocket2!15} 0.83 $\pm$ 0.02
  & 1.47 $\pm$ 0.06 & \cellcolor{rocket2!15} 0.12 $\pm$ 0.01 & 0.92 $\pm$ 0.01
  & 2.59 $\pm$ 0.11 & 0.35 $\pm$ 0.01 & 0.75 $\pm$ 0.01 \\

\hdashline[4pt/2pt]
\noalign{\vskip 0.3ex}

\textsf{\footnotesize CF\textsubscript{Dif.}}
  & \cellcolor{rocket2!15} 4.43 $\pm$ 0.29 & \cellcolor{rocket2!15} 0.23 $\pm$ 0.02 & \cellcolor{rocket2!30}0.75 $\pm$ 0.02 
  & \cellcolor{rocket2!60}\textbf{1.31 $\pm$ 0.07} & \cellcolor{rocket2!60}\textbf{0.15 $\pm$ 0.01} & \cellcolor{rocket2!60}\textbf{0.85 $\pm$ 0.01}
  & \cellcolor{rocket2!15}2.49 $\pm$ 0.12 & \cellcolor{rocket2!15}0.44 $\pm$ 0.01 & \cellcolor{rocket2!60}\textbf{0.60 $\pm$ 0.01} \\

\textsf{\footnotesize EG (CF)\textsubscript{Dif.}}
  & \cellcolor{rocket2!30} 4.30 $\pm$ 0.29 & \cellcolor{rocket2!30} 0.26 $\pm$ 0.02 & \cellcolor{rocket2!45}0.74 $\pm$ 0.02 
  & \cellcolor{rocket2!60}\textbf{1.31 $\pm$ 0.07} & \cellcolor{rocket2!60}\textbf{0.15 $\pm$ 0.01} & \cellcolor{rocket2!60}\textbf{0.85 $\pm$ 0.01$^{*}$}
  & \cellcolor{rocket2!30}2.48 $\pm$ 0.12 & \cellcolor{rocket2!45}0.45 $\pm$ 0.01 & \cellcolor{rocket2!60}\textbf{0.60 $\pm$ 0.01$^{*}$} \\

\hdashline[2pt/2pt]
\noalign{\vskip 0.3ex}

\textsf{\footnotesize CF\textsubscript{VAE}}
  & \cellcolor{rocket2!60}\textbf{4.16 $\pm$ 0.31$^{*}$} & \cellcolor{rocket2!60}\textbf{0.32 $\pm$ 0.02$^{*}$} & \cellcolor{rocket2!60}\textbf{0.72 $\pm$ 0.02$^{*}$}
  & \cellcolor{rocket2!30} 1.34 $\pm$ 0.07 & \cellcolor{rocket2!30} 0.13 $\pm$ 0.01 & \cellcolor{rocket2!30} 0.88 $\pm$ 0.01
  & \cellcolor{rocket2!45}2.45 $\pm$ 0.12 & \cellcolor{rocket2!45}0.45 $\pm$ 0.01 & \cellcolor{rocket2!45}0.61 $\pm$ 0.01 \\

\textsf{\footnotesize EG (CF)\textsubscript{VAE}}
  & \cellcolor{rocket2!45}4.17 $\pm$ 0.30 & \cellcolor{rocket2!45}0.31 $\pm$ 0.02 & \cellcolor{rocket2!45}0.74 $\pm$ 0.02 & \cellcolor{rocket2!60}\textbf{1.31 $\pm$ 0.07} & \cellcolor{rocket2!45}0.14 $\pm$ 0.01 & \cellcolor{rocket2!45}0.87 $\pm$ 0.01
  & \cellcolor{rocket2!60}\textbf{2.42 $\pm$ 0.12$^{*}$} & \cellcolor{rocket2!60}\textbf{0.46 $\pm$ 0.01$^{*}$} & \cellcolor{rocket2!30}0.62 $\pm$ 0.01 \\
   
    \bottomrule
    \end{tabular}
  \label{tab:full_results}
  \vspace{-1em}
\end{table*}

\section{Results}

\begin{figure}[b]
\vspace{-4mm}
    \centering
    \includegraphics[width=0.9\linewidth]{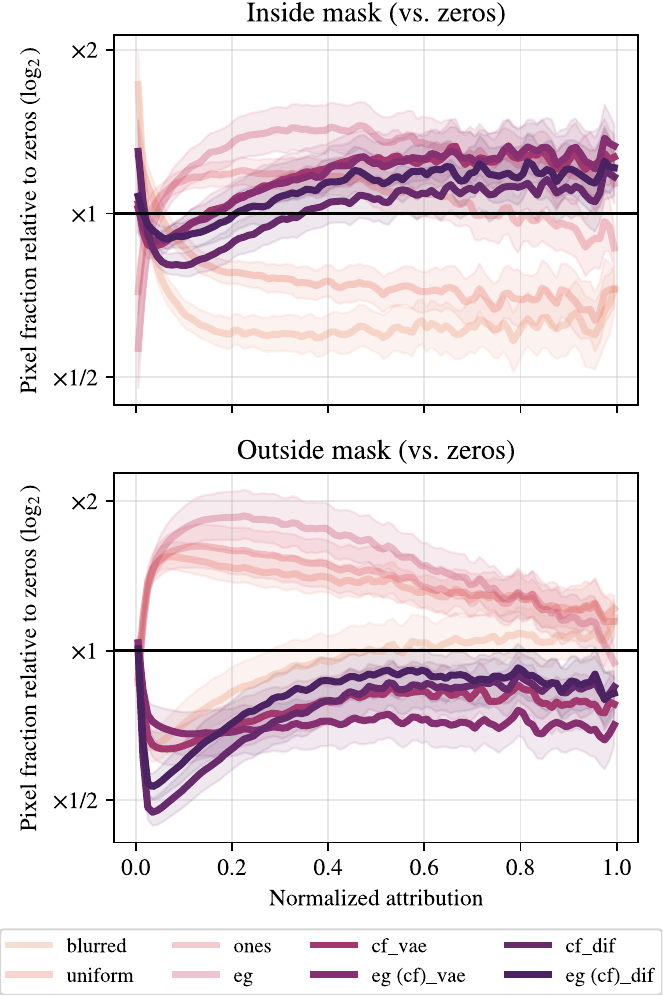}
    \caption{
    The distributions of relative pixel fractions compared to the zeros baseline, computed over 240 samples (80 samples per dataset) (mean $\pm$ 95\% confidence interval).
    }
    \label{fig:attribution_distribution_main}
\end{figure}

\textbf{Comparison to standard baselines} 
Fig.~\ref{fig:attribution_distribution_main} shows the distributions of attribution intensities relative to the \textsf{\small Zeros} baseline. Measured across all three data sets, the counterfactual baselines clearly yield a larger fraction of high-intensity attributions within the ground-truth mask and a smaller fraction outside of it. These results indicate that counterfactual baselines produce attribution maps that concentrate more strongly on the annotated regions and contain fewer spurious attributions in the background. This finding is consistent with our qualitative analysis, where the proposed baselines yield cleaner maps with fewer spurious attributions, as illustrated in Fig.~\ref{fig:examples} and Appendix~\ref{appendix_sec_examples}.

The mask-overlap metrics, computed separately for each data set (Table~\ref{tab:full_results}), confirm these observations. We assess statistical significance between the best-performing counterfactual baseline and the best non-counterfactual baseline using paired $t$-tests or Wilcoxon signed-rank tests, depending on the normality of the paired differences. The counterfactual baselines achieve the highest or among the highest scores across all data sets and metrics; only on the Chest X-ray data set do \textsf{\small Ones} and \textsf{\small Blurred} perform comparably. The FPAR metric is particularly relevant in the medical setting, as false positive attributions may direct attention to clinically irrelevant regions. Here, the counterfactual baselines substantially reduce false positive attributions on all data sets compared with the best non-counterfactual baseline. Relative to the most commonly used \textsf{\small Zeros} baseline, this corresponds to a reduction of up to 15\% on the manometry data set and up to 8\% on the Chest X-ray and brain data sets, achieved solely by changing the baseline, without modifying the attribution method or the classifier.

Taken together, the quantitative and qualitative results show a clear and consistent advantage of counterfactual baselines. While individual standard baselines are competitive only in specific settings, e.g., on a single data set or metric, the counterfactual baselines rank among the best across all data sets and evaluation criteria, indicating that their benefit generalizes across the diverse medical domains studied.

\textbf{Comparison to pure counterfactual explanations}
\setlength\dashlinedash{1pt}
\setlength\dashlinegap{1pt}
\setlength\arrayrulewidth{0.6pt}
\begin{table*}
  \footnotesize
  \centering
  \caption{Aggregated scores (mean $\pm$ se) comparing proposed baselines to related explainability approaches; p $<$ 0.01 (*)}
   \setlength{\tabcolsep}{1.8pt}

    \begin{tabular}{
  l
  >{\hskip 0.04em}c<{\hskip 0.04em}
  >{\hskip 0.04em}c<{\hskip 0.04em}
  >{\hskip 0.04em}c<{\hskip 0.04em}
  >{\hskip 0.04em}c<{\hskip 0.04em}
  >{\hskip 0.04em}c<{\hskip 0.04em}
  >{\hskip 0.04em}c<{\hskip 0.04em}
  >{\hskip 0.04em}c<{\hskip 0.04em}
  >{\hskip 0.04em}c<{\hskip 0.04em}
  >{\hskip 0.04em}c<{\hskip 0.04em}
}
    \toprule
     & \multicolumn{3}{c}{\textbf{Manometry}} & \multicolumn{3}{c}{\textbf{Chest X-ray}} & \multicolumn{3}{c}{\textbf{Brain MRI}} \\
    \cmidrule(lr){2-4} \cmidrule(lr){5-7} \cmidrule(lr){8-10}
    \textbf{Baseline} & \multicolumn{1}{c}{MSE $\downarrow$ {\fontsize{4}{5}\selectfont{$\left(\cdot10^{\text{-}2}\right)$}}} & \multicolumn{1}{c}{PR-AUC $\uparrow$} & \multicolumn{1}{c}{FPAR $\downarrow$} & \multicolumn{1}{c}{MSE $\downarrow$ {\fontsize{4}{5}\selectfont{$\left(\cdot10^{\text{-}2}\right)$}}} & \multicolumn{1}{c}{PR-AUC $\uparrow$} & \multicolumn{1}{c}{FPAR $\downarrow$} & \multicolumn{1}{c}{MSE $\downarrow$ {\fontsize{4}{5}\selectfont{$\left(\cdot10^{\text{-}2}\right)$}}} & \multicolumn{1}{c}{PR-AUC $\uparrow$} & \multicolumn{1}{c}{FPAR $\downarrow$} \\
    \midrule
\textsf{\footnotesize DIFF\textsubscript{Dif.}} & 5.87 $\pm$ 0.24 & 0.14 $\pm$ 0.01  & 0.91 $\pm$ 0.01 & 3.21 $\pm$ 0.09 & 0.03 $\pm$ 0.00  & 0.97 $\pm$ 0.00 & 4.28 $\pm$ 0.09 & 0.14 $\pm$ 0.01 & 0.90 $\pm$ 0.00  \\    
\textsf{\footnotesize DIFF\textsubscript{VAE}} & 5.88 $\pm$ 0.32 & \cellcolor{rocket2!60}\textbf{0.33 $\pm$ 0.02} & 0.87 $\pm$ 0.01 & 2.58 $\pm$ 0.08 & 0.03 $\pm$ 0.00 & 0.97 $\pm$ 0.00 & 4.22 $\pm$ 0.09 & 0.16 $\pm$ 0.01 & 0.89 $\pm$ 0.00 \\  

\hdashline[2pt/2pt]
\noalign{\vskip 0.3ex}

\textsf{\footnotesize Mean (n)} & 4.46 $\pm$ 0.26 & 0.21 $\pm$ 0.02 & 0.81 $\pm$ 0.01 & \cellcolor{rocket2!15} 1.44 $\pm$ 0.07 & 0.11 $\pm$ 0.01 & 0.91 $\pm$ 0.01 & 2.95 $\pm$ 0.12 & 0.18 $\pm$ 0.01 & 0.83 $\pm$ 0.01 \\
\textsf{\footnotesize Rand (n)} & 4.57 $\pm$ 0.28 & 0.18 $\pm$ 0.02 & 0.83 $\pm$ 0.01 & 1.60 $\pm$ 0.07 & 0.07 $\pm$ 0.01 & 0.94 $\pm$ 0.00 & 2.70 $\pm$ 0.12 & 0.28 $\pm$ 0.01 & 0.75 $\pm$ 0.01 \\
\textsf{\footnotesize EG (n)} & 4.40 $\pm$ 0.25 & 0.24 $\pm$ 0.02 & 0.82 $\pm$ 0.01 & 1.48 $\pm$ 0.07 & \cellcolor{rocket2!15} 0.12 $\pm$ 0.01 & 0.92 $\pm$ 0.00 & 2.54 $\pm$ 0.11 & 0.37 $\pm$ 0.01 & 0.74 $\pm$ 0.01 \\
\textsf{\footnotesize LIG} \cite{dravidMedXGANVisualExplanations2022} & \cellcolor{rocket2!15} 4.37 $\pm$ 0.29 & 0.24 $\pm$ 0.02& 0.80 $\pm$ 0.02 & 1.52 $\pm$ 0.07 & 0.04 $\pm$ 0.00 & 0.96 $\pm$ 0.00 & 3.09 $\pm$ 0.12 & 0.15 $\pm$ 0.01 & 0.84 $\pm$ 0.01 \\
\textsf{\footnotesize CF-IG} \cite{duellCounterfactualIntegratedGradientsCounterfactual2023} &  4.43 $\pm$ 0.28  & 0.20 $\pm$ 0.01  &  0.81 $\pm$ 0.01  & \cellcolor{rocket2!30} 1.43 $\pm$ 0.07 & \cellcolor{rocket2!15} 0.12 $\pm$ 0.01  & \cellcolor{rocket2!15} 0.90 $\pm$ 0.01 & 2.72 $\pm$ 0.13 & 0.31 $\pm$ 0.01 & 0.72 $\pm$ 0.01  \\   
\textsf{\footnotesize GIG \cite{kapishnikovGuidedIntegratedGradients2021}} & 4.50 $\pm$ 0.29 &  0.19 $\pm$ 0.02 & 0.85 $\pm$ 0.01 & 1.52 $\pm$ 0.06 & \cellcolor{rocket2!15} 0.12 $\pm$ 0.01 & 0.92 $\pm$ 0.01 & 2.51 $\pm$ 0.12 & 0.39 $\pm$ 0.01 & 0.65 $\pm$ 0.01 \\  
\textsf{\footnotesize IG2 \cite{zhuoIG2IntegratedGradient2024a}} &  4.76 $\pm$ 0.30  & 0.11 $\pm$ 0.01 & 0.88 $\pm$ 0.01 & 1.55 $\pm$ 0.08 & 0.04 $\pm$ 0.00  & 0.96 $\pm$ 0.00 & 3.17 $\pm$ 0.13 & 0.07 $\pm$ 0.00 & 0.91 $\pm$ 0.00 \\  
\hdashline[4pt/2pt]
\noalign{\vskip 0.3ex}

\textsf{\footnotesize CF\textsubscript{Dif.}}
  & 4.43 $\pm$ 0.29 & 0.23 $\pm$ 0.02 & \cellcolor{rocket2!30}0.75 $\pm$ 0.02 
  & \cellcolor{rocket2!60}\textbf{1.31 $\pm$ 0.07} & \cellcolor{rocket2!60}\textbf{0.15 $\pm$ 0.01} & \cellcolor{rocket2!60}\textbf{0.85 $\pm$ 0.01}
  & \cellcolor{rocket2!15}2.49 $\pm$ 0.12 & \cellcolor{rocket2!15}0.44 $\pm$ 0.01 & \cellcolor{rocket2!60}\textbf{0.60 $\pm$ 0.01} \\

\textsf{\footnotesize EG (CF)\textsubscript{Dif.}}
  & \cellcolor{rocket2!30} 4.30 $\pm$ 0.29 & \cellcolor{rocket2!15} 0.26 $\pm$ 0.02 & \cellcolor{rocket2!45}0.74 $\pm$ 0.02 
  & \cellcolor{rocket2!60}\textbf{1.31 $\pm$ 0.07$^{*}$} & \cellcolor{rocket2!60}\textbf{0.15 $\pm$ 0.01$^{*}$} & \cellcolor{rocket2!60}\textbf{0.85 $\pm$ 0.01$^{*}$}
  & \cellcolor{rocket2!30}2.48 $\pm$ 0.12 & \cellcolor{rocket2!45}0.45 $\pm$ 0.01 & \cellcolor{rocket2!60}\textbf{0.60 $\pm$ 0.01$^{*}$} \\

\hdashline[2pt/2pt]
\noalign{\vskip 0.3ex}

\textsf{\footnotesize CF\textsubscript{VAE}}
  & \cellcolor{rocket2!60}\textbf{4.16 $\pm$ 0.31$^{*}$} & \cellcolor{rocket2!45}0.32 $\pm$ 0.02 & \cellcolor{rocket2!60}\textbf{0.72 $\pm$ 0.02$^{*}$}
  & \cellcolor{rocket2!45} 1.34 $\pm$ 0.07 & \cellcolor{rocket2!30} 0.13 $\pm$ 0.01 & \cellcolor{rocket2!30} 0.88 $\pm$ 0.01
  & \cellcolor{rocket2!45}2.45 $\pm$ 0.12 & \cellcolor{rocket2!45}0.45 $\pm$ 0.01 & \cellcolor{rocket2!45}0.61 $\pm$ 0.01 \\

\textsf{\footnotesize EG (CF)\textsubscript{VAE}}
  & \cellcolor{rocket2!45}4.17 $\pm$ 0.30 & \cellcolor{rocket2!30}0.31 $\pm$ 0.02 & \cellcolor{rocket2!45}0.74 $\pm$ 0.02 & \cellcolor{rocket2!60}\textbf{1.31 $\pm$ 0.07} & \cellcolor{rocket2!45}0.14 $\pm$ 0.01 & \cellcolor{rocket2!45}0.87 $\pm$ 0.01
  & \cellcolor{rocket2!60}\textbf{2.42 $\pm$ 0.12$^{*}$} & \cellcolor{rocket2!60}\textbf{0.46 $\pm$ 0.01$^{*}$} & \cellcolor{rocket2!30}0.62 $\pm$ 0.01 \\
  
    \bottomrule
    \end{tabular}
  \label{tab:other_results}
 \vspace{-1.5em}
\end{table*}
Beyond their role as baselines, counterfactuals can also serve directly as explanations themselves, which is a separate explainability paradigm in its own. We take the difference $(x - \hat{\text{CF}})$ as the attribution map and then compare it to the attributions obtained with our proposed baseline approaches. While this raw difference can offer intuitive and visually informative insights, it lacks quantitative interpretability and its quality degrades rapidly when the counterfactual is imperfect, as the raw difference then captures spurious variation across the entire image rather than localizing to the pathologically relevant region. Using the counterfactual as a baseline for IG compensates for this by weighting feature contributions according to model sensitivity along the path, rather than pixel-wise distance alone. It also satisfies key axiomatic properties, which ensure theoretical soundness and consistence. A simple difference map does not guarantee those and therefore lacks the robustness required for reliable interpretability. 

\begin{figure}[b]
\vspace{-4mm}
\centering
\includegraphics[width=0.75\linewidth]{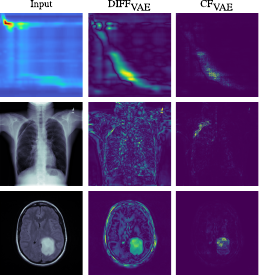}
\caption{
Comparison of the raw difference between the counterfactual and input (\textsf{\small DIFF\textsubscript{VAE}}) vs. using IG with {\small CF\textsubscript{VAE}} baseline
}
\label{fig:diff_comparison}
\vspace{-1em}
\end{figure}

This is reflected in our results: the absolute difference maps (\textsf{\small DIFF\textsubscript{Dif.}} and \textsf{\small DIFF\textsubscript{VAE}}) in Table~\ref{tab:other_results} perform worse than the proposed baselines across all metrics except PR-AUC on the manometry data set. Specifically the FPAR metric clearly shows that the pure difference is highlighting substantially more false positive regions outside the medically important region of interest. Qualitative comparisons also support this finding. Fig.~\ref{fig:diff_comparison} shows an example, demonstrating how \textsf{\small CF\textsubscript{VAE}} attributions concentrate precisely on the region of interest while \textsf{\small DIFF\textsubscript{VAE}} highlights diffuse regions throughout the image. These findings reinforce the value of counterfactuals as principled baselines for IG over their use as standalone explanations. An extended discussion and additional qualitative results are provided in Appendix~\ref{appendix_difference}.

\textbf{Comparison to related concepts}
Our approach is conceptually distinct, yet grounded in similar underlying principles, thus we compare it to the related methods LIG~\cite{dravidMedXGANVisualExplanations2022} and CF-IG~\cite{duellCounterfactualIntegratedGradientsCounterfactual2023}, as well as to the adaptive path methods GIG~\cite{kapishnikovGuidedIntegratedGradients2021}, and IG2~\cite{zhuoIG2IntegratedGradient2024a}.
Additionally, our analysis also concerns the necessity of generating a dedicated counterfactual for each input, as an alternative and potentially simpler strategy might be to use samples from the normal class as baselines for IG. To investigate this, we additionally evaluate three baseline strategies that sample directly from the normal class in the training set:
\textsf{\small Mean (n)}: the mean of all normal samples;
\textsf{\small Rand (n)}: a single random normal sample;
\textsf{\small EG (n)}: EG with 50 random normal samples.

The results in Table~\ref{tab:other_results} show that the counterfactual baselines consistently outperform those approaches. Again, the consistency is particularly noteworthy: none of the other approaches achieves good scores across all metrics and data sets. These results highlight the benefit of input-specific baselines. Additional qualitative results and implementation details are provided in Appendix~\ref{sec:appendix_otherstrategies}.
\section{Conclusion}
In this work we analyze counterfactual baselines for path attribution methods. Addressing the limitations of standard baselines, we introduce \emph{semantic missingness} as a stricter requirement for baselines in medical settings, which motivates clinically ``normal'' counterfactuals as baselines. Theoretical analysis and experiments on three medical data sets show that counterfactual baselines yield more localized and clinically relevant attributions than conventional baselines and related methods, and that using counterfactuals as IG baselines outperforms using them directly as explanations, bridging two complementary explainability paradigms.

\textbf{Limitations}
The main limitation of our approach is the additional computational cost, which we discuss in Appendix~\ref{sec:limitations}.

\textbf{Future work}
As our framework is model-agnostic, more advanced generative models could improve counterfactual fidelity and thus attribution quality. Stronger classifiers and ensembles of counterfactual and conventional baselines may increase the trustworthiness and robustness of explanations, toward more clinically meaningful explainability. Beyond healthcare, our framework offers a general strategy for domain-aware explanations in other safety-critical settings.
{
    \small
    \bibliographystyle{ieeenat_fullname}
    \bibliography{bibliography}

\begin{thebibliography}{52}
\providecommand{\natexlab}[1]{#1}
\providecommand{\url}[1]{\texttt{#1}}
\expandafter\ifx\csname urlstyle\endcsname\relax
  \providecommand{\doi}[1]{doi: #1}\else
  \providecommand{\doi}{doi: \begingroup \urlstyle{rm}\Url}\fi

\bibitem[Adebayo et~al.(2018)Adebayo, Gilmer, Muelly, Goodfellow, Hardt, and
  Kim]{adebayoSanityChecksSaliency2018}
Julius Adebayo, Justin Gilmer, Michael Muelly, Ian Goodfellow, Moritz Hardt,
  and Been Kim.
\newblock Sanity {{Checks}} for {{Saliency Maps}}.
\newblock In \emph{Advances in {{Neural Information Processing Systems}}}.
  Curran Associates, Inc., 2018.

\bibitem[Ancona et~al.(2018)Ancona, Ceolini, {\"O}ztireli, and
  Gross]{anconaBetterUnderstandingGradientbased2018a}
Marco Ancona, Enea Ceolini, Cengiz {\"O}ztireli, and Markus Gross.
\newblock Towards better understanding of gradient-based attribution methods
  for {{Deep Neural Networks}}.
\newblock In \emph{6th {{International Conference}} on {{Learning
  Representations}}, {{ICLR}} 2018}, 2018.

\bibitem[Angelov et~al.(2021)Angelov, Soares, Jiang, Arnold, and
  Atkinson]{angelovExplainableArtificialIntelligence2021}
Plamen~P. Angelov, Eduardo~A. Soares, Richard Jiang, Nicholas~I. Arnold, and
  Peter~M. Atkinson.
\newblock Explainable artificial intelligence: An analytical review.
\newblock \emph{WIREs Data Mining and Knowledge Discovery}, 11\penalty0
  (5):\penalty0 e1424, 2021.

\bibitem[Arras et~al.(2022)Arras, Osman, and
  Samek]{arrasCLEVRXAIBenchmarkDataset2022}
Leila Arras, Ahmed Osman, and Wojciech Samek.
\newblock {{CLEVR-XAI}}: {{A}} benchmark dataset for the ground truth
  evaluation of neural network explanations.
\newblock \emph{Information Fusion}, 81:\penalty0 14--40, 2022.

\bibitem[Atad et~al.(2024)Atad, Schinz, Moeller, Graf, Wiestler, Rueckert,
  Navab, Kirschke, and Keicher]{atadCounterfactualExplanationsMedical2024}
Matan Atad, David Schinz, Hendrik Moeller, Robert Graf, Benedikt Wiestler,
  Daniel Rueckert, Nassir Navab, Jan~S. Kirschke, and Matthias Keicher.
\newblock Counterfactual {{Explanations}} for {{Medical Image Classification}}
  and {{Regression}} using {{Diffusion Autoencoder}}.
\newblock \emph{Machine Learning for Biomedical Imaging}, 2\penalty0 (iMIMIC
  2023 special issue):\penalty0 2103--2125, 2024.

\bibitem[Bass et~al.(2020)Bass, {da Silva}, Sudre, Tudosiu, Smith, and
  Robinson]{bassICAMInterpretableClassification2020}
Cher Bass, Mariana {da Silva}, Carole Sudre, Petru-Daniel Tudosiu, Stephen
  Smith, and Emma Robinson.
\newblock {{ICAM}}: {{Interpretable Classification}} via {{Disentangled
  Representations}} and {{Feature Attribution Mapping}}.
\newblock In \emph{Advances in {{Neural Information Processing Systems}}},
  pages 7697--7709. Curran Associates, Inc., 2020.

\bibitem[Baumgartner et~al.(2018)Baumgartner, Koch, Tezcan, Ang, and
  Konukoglu]{baumgartnerVisualFeatureAttribution2018}
Christian~F. Baumgartner, Lisa~M. Koch, Kerem~Can Tezcan, Jia~Xi Ang, and Ender
  Konukoglu.
\newblock Visual {{Feature Attribution Using Wasserstein GANs}}.
\newblock In \emph{2018 {{IEEE}}/{{CVF Conference}} on {{Computer Vision}} and
  {{Pattern Recognition}}}, pages 8309--8319, Salt Lake City, UT, USA, 2018.
  IEEE.

\bibitem[Brandenburg et~al.(2025)Brandenburg, {M{\"u}ller-Stich}, Wagner, and
  van~der Schaar]{brandenburgCanSurgeonsTrust2025}
Johanna~M. Brandenburg, Beat~P. {M{\"u}ller-Stich}, Martin Wagner, and Mihaela
  van~der Schaar.
\newblock Can surgeons trust {{AI}}? {{Perspectives}} on machine learning in
  surgery and the importance of {{eXplainable Artificial Intelligence}}
  ({{XAI}}).
\newblock \emph{Langenbeck's Archives of Surgery}, 410\penalty0 (1), 2025.

\bibitem[Buda et~al.(2019)Buda, Saha, and
  Mazurowski]{budaAssociationGenomicSubtypes2019}
Mateusz Buda, Ashirbani Saha, and Maciej Mazurowski.
\newblock Association of genomic subtypes of lower-grade gliomas with shape
  features automatically extracted by a deep learning algorithm.
\newblock \emph{Computers in Biology and Medicine}, 109, 2019.

\bibitem[Bylinskii et~al.(2019)Bylinskii, Judd, Oliva, Torralba, and
  Durand]{bylinskiiWhatDifferentEvaluation2019}
Zoya Bylinskii, Tilke Judd, Aude Oliva, Antonio Torralba, and Fredo Durand.
\newblock What {{Do Different Evaluation Metrics Tell Us About Saliency
  Models}}?
\newblock \emph{IEEE Transactions on Pattern Analysis and Machine
  Intelligence}, 41\penalty0 (3):\penalty0 740--757, 2019.

\bibitem[Chaddad et~al.(2023)Chaddad, Peng, Xu, and
  Bouridane]{chaddadSurveyExplainableAI2023}
Ahmad Chaddad, Jihao Peng, Jian Xu, and Ahmed Bouridane.
\newblock Survey of {{Explainable AI Techniques}} in {{Healthcare}}.
\newblock \emph{Sensors}, 23\penalty0 (2):\penalty0 634, 2023.

\bibitem[Clough et~al.(2019)Clough, Oksuz, {Puyol-Ant{\'o}n}, Ruijsink, King,
  and Schnabel]{cloughGlobalLocalInterpretability2019}
James~R. Clough, Ilkay Oksuz, Esther {Puyol-Ant{\'o}n}, Bram Ruijsink,
  Andrew~P. King, and Julia~A. Schnabel.
\newblock Global and {{Local Interpretability}} for {{Cardiac MRI
  Classification}}.
\newblock In \emph{Medical {{Image Computing}} and {{Computer Assisted
  Intervention}} -- {{MICCAI}} 2019}, pages 656--664, Cham, 2019. Springer
  International Publishing.

\bibitem[Dabkowski and Gal(2017)]{dabkowskiRealTimeImage2017}
Piotr Dabkowski and Yarin Gal.
\newblock Real {{Time Image Saliency}} for {{Black Box Classifiers}}.
\newblock In \emph{Advances in {{Neural Information Processing Systems}}}.
  Curran Associates, Inc., 2017.

\bibitem[Di~Folco et~al.(2024)Di~Folco, Bercea, Chan, and
  Schnabel]{difolcoInterpretableRepresentationLearning2024}
Maxime Di~Folco, Cosmin~I. Bercea, Emily Chan, and Julia~A. Schnabel.
\newblock Interpretable {{Representation Learning}} of~{{Cardiac MRI}}
  via~{{Attribute Regularization}}.
\newblock In \emph{Medical {{Image Computing}} and {{Computer Assisted
  Intervention}} -- {{MICCAI}} 2024}, pages 492--501, Cham, 2024. Springer
  Nature Switzerland.

\bibitem[Dravid et~al.(2022)Dravid, Schiffers, Gong, and
  Katsaggelos]{dravidMedXGANVisualExplanations2022}
Amil Dravid, Florian Schiffers, Boqing Gong, and Aggelos~K. Katsaggelos.
\newblock {{medXGAN}}: {{Visual Explanations}} for {{Medical Classifiers}}
  through a {{Generative Latent Space}}.
\newblock In \emph{2022 {{IEEE}}/{{CVF Conference}} on {{Computer Vision}} and
  {{Pattern Recognition Workshops}} ({{CVPRW}})}, pages 2935--2944, New
  Orleans, LA, USA, 2022. IEEE.

\bibitem[Duell et~al.(2023)Duell, Seisenberger, and
  Fan]{duellCounterfactualIntegratedGradientsCounterfactual2023}
Jamie Duell, Monika Seisenberger, and Xiuyi Fan.
\newblock Counterfactual-{{Integrated Gradients}}: {{Counterfactual Feature
  Attribution}} for {{Medical Records}}.
\newblock In \emph{2023 {{IEEE International Conference}} on {{Bioinformatics}}
  and {{Biomedicine}}}, 2023.

\bibitem[Erion et~al.(2021)Erion, Janizek, Sturmfels, Lundberg, and
  Lee]{erionImprovingPerformanceDeep2021}
Gabriel Erion, Joseph~D. Janizek, Pascal Sturmfels, Scott~M. Lundberg, and
  Su-In Lee.
\newblock Improving performance of deep learning models with axiomatic
  attribution priors and expected gradients.
\newblock \emph{Nature Machine Intelligence}, 3\penalty0 (7):\penalty0
  620--631, 2021.

\bibitem[Fong and Vedaldi(2017)]{fongInterpretableExplanationsBlack2017}
Ruth~C. Fong and Andrea Vedaldi.
\newblock Interpretable {{Explanations}} of {{Black Boxes}} by {{Meaningful
  Perturbation}}.
\newblock In \emph{2017 {{IEEE International Conference}} on {{Computer
  Vision}} ({{ICCV}})}, pages 3449--3457, Venice, 2017. IEEE.

\bibitem[Ghorbani et~al.(2019)Ghorbani, Abid, and
  Zou]{ghorbaniInterpretationNeuralNetworks2019}
Amirata Ghorbani, Abubakar Abid, and James Zou.
\newblock Interpretation of {{Neural Networks Is Fragile}}.
\newblock \emph{Proceedings of the AAAI Conference on Artificial Intelligence},
  33\penalty0 (01):\penalty0 3681--3688, 2019.

\bibitem[Guyomard et~al.(2022)Guyomard, Fessant, Guyet, {termier}, and
  Bouadi]{guyomardVCNetSelfexplainingModel2022}
Victor Guyomard, Fran{\c c}oise Fessant, Thomas Guyet, Alexandre {termier}, and
  Tassadit Bouadi.
\newblock {{VCNet}}: {{A}} self-explaining model for realistic counterfactual
  generation.
\newblock In \emph{Machine {{Learning}} and {{Knowledge Discovery}} in
  {{Databases}} - {{European Conference}}, {{ECML PKDD}} 2022, {{Grenoble}},
  {{France}}, {{September}} 19-23, 2022, {{Proceedings}}, {{Part I}}}, 2022.

\bibitem[Ho and Salimans(2022)]{hoClassifierFreeDiffusionGuidance2022}
Jonathan Ho and Tim Salimans.
\newblock Classifier-{{Free Diffusion Guidance}}, 2022.

\bibitem[Ho et~al.(2020)Ho, Jain, and
  Abbeel]{hoDenoisingDiffusionProbabilistic2020}
Jonathan Ho, Ajay Jain, and Pieter Abbeel.
\newblock Denoising {{Diffusion Probabilistic Models}}.
\newblock In \emph{Advances in {{Neural Information Processing Systems}}},
  pages 6840--6851. Curran Associates, Inc., 2020.

\bibitem[Hooker et~al.(2019)Hooker, Erhan, Kindermans, and
  Kim]{hookerBenchmarkInterpretabilityMethods2019}
Sara Hooker, Dumitru Erhan, Pieter-Jan Kindermans, and Been Kim.
\newblock A {{Benchmark}} for {{Interpretability Methods}} in {{Deep Neural
  Networks}}.
\newblock In \emph{Advances in {{Neural Information Processing Systems}}}.
  Curran Associates, Inc., 2019.

\bibitem[Hossain et~al.(2025)Hossain, Zamzmi, Mouton, Salekin, Sun, and
  Goldgof]{hossainExplainableAIMedical2025a}
Md~Imran Hossain, Ghada Zamzmi, Peter~R. Mouton, Md~Sirajus Salekin, Yu Sun,
  and Dmitry Goldgof.
\newblock Explainable {{AI}} for {{Medical Data}}: {{Current Methods}},
  {{Limitations}}, and {{Future Directions}}.
\newblock \emph{ACM Computing Surveys}, 57\penalty0 (6):\penalty0 1--46, 2025.

\bibitem[Kadir et~al.(2023)Kadir, Mosavi, and
  Sonntag]{kadirEvaluationMetricsXAI2023}
Md~Abdul Kadir, Amir Mosavi, and Daniel Sonntag.
\newblock Evaluation {{Metrics}} for {{XAI}}: {{A Review}}, {{Taxonomy}}, and
  {{Practical Applications}}.
\newblock In \emph{2023 {{IEEE}} 27th {{International Conference}} on
  {{Intelligent Engineering Systems}} ({{INES}})}, pages 000111--000124,
  Nairobi, Kenya, 2023. IEEE.

\bibitem[Kapishnikov et~al.(2021)Kapishnikov, Venugopalan, Avci, Wedin, Terry,
  and Bolukbasi]{kapishnikovGuidedIntegratedGradients2021}
Andrei Kapishnikov, Subhashini Venugopalan, Besim Avci, Ben Wedin, Michael
  Terry, and Tolga Bolukbasi.
\newblock Guided {{Integrated Gradients}}: An {{Adaptive Path Method}} for
  {{Removing Noise}}.
\newblock In \emph{2021 {{IEEE}}/{{CVF Conference}} on {{Computer Vision}} and
  {{Pattern Recognition}} ({{CVPR}})}, pages 5048--5056, Nashville, TN, USA,
  2021. IEEE.

\bibitem[Kim et~al.(2022)Kim, Meister, Ramaswamy, Fong, and
  Russakovsky]{kimHIVEEvaluatingHuman2022}
Sunnie S.~Y. Kim, Nicole Meister, Vikram~V. Ramaswamy, Ruth Fong, and Olga
  Russakovsky.
\newblock {{HIVE}}: {{Evaluating}} the {{Human Interpretability}} of {{Visual
  Explanations}}.
\newblock In \emph{Computer {{Vision}} -- {{ECCV}} 2022}, pages 280--298, Cham,
  2022. Springer Nature Switzerland.

\bibitem[Kindermans et~al.(2019)Kindermans, Hooker, Adebayo, Alber, Sch{\"u}tt,
  D{\"a}hne, Erhan, and Kim]{kindermansUnreliabilitySaliencyMethods2019}
Pieter-Jan Kindermans, Sara Hooker, Julius Adebayo, Maximilian Alber,
  Kristof~T. Sch{\"u}tt, Sven D{\"a}hne, Dumitru Erhan, and Been Kim.
\newblock The ({{Un}})reliability of {{Saliency Methods}}.
\newblock In \emph{Explainable {{AI}}: {{Interpreting}}, {{Explaining}} and
  {{Visualizing Deep Learning}}}, pages 267--280. Springer International
  Publishing, Cham, 2019.

\bibitem[Kingma and Welling(2014)]{kingmaAutoEncodingVariationalBayes2014}
Diederik~P. Kingma and Max Welling.
\newblock Auto-{{Encoding Variational Bayes}}.
\newblock In \emph{2nd {{International Conference}} on {{Learning
  Representations}}, {{ICLR}} 2014, {{Banff}}, {{AB}}, {{Canada}}, {{April}}
  14-16, 2014, {{Conference Track Proceedings}}}, 2014.

\bibitem[Lage et~al.(2019)Lage, Chen, He, Narayanan, Kim, Gershman, and
  {Doshi-Velez}]{lageEvaluationHumanInterpretabilityExplanation2019}
Isaac Lage, Emily Chen, Jeffrey He, Menaka Narayanan, Been Kim, Sam Gershman,
  and Finale {Doshi-Velez}.
\newblock An {{Evaluation}} of the {{Human-Interpretability}} of
  {{Explanation}}, 2019.

\bibitem[Liu et~al.(2022)Liu, Hu, Lin, Yao, Xie, Wei, Ning, Cao, Zhang, Dong,
  Wei, and Guo]{liuSwinTransformerV22022}
Ze Liu, Han Hu, Yutong Lin, Zhuliang Yao, Zhenda Xie, Yixuan Wei, Jia Ning, Yue
  Cao, Zheng Zhang, Li Dong, Furu Wei, and Baining Guo.
\newblock Swin {{Transformer V2}}: {{Scaling Up Capacity}} and {{Resolution}}.
\newblock In \emph{2022 {{IEEE}}/{{CVF Conference}} on {{Computer Vision}} and
  {{Pattern Recognition}} ({{CVPR}})}, pages 11999--12009, 2022.

\bibitem[Lundberg and Lee(2017)]{lundbergUnifiedApproachInterpreting2017}
Scott~M. Lundberg and Su-In Lee.
\newblock A unified approach to interpreting model predictions.
\newblock In \emph{Proceedings of the 31st {{International Conference}} on
  {{Neural Information Processing Systems}}}, pages 4768--4777, Red Hook, NY,
  USA, 2017. Curran Associates Inc.

\bibitem[Lundstrom et~al.(2022)Lundstrom, Huang, and
  Razaviyayn]{lundstromRigorousStudyIntegrated2022}
Daniel~D. Lundstrom, Tianjian Huang, and Meisam Razaviyayn.
\newblock A {{Rigorous Study}} of {{Integrated Gradients Method}} and
  {{Extensions}} to {{Internal Neuron Attributions}}.
\newblock In \emph{Proceedings of the 39th {{International Conference}} on
  {{Machine Learning}}}, pages 14485--14508. PMLR, 2022.

\bibitem[Mertes et~al.(2022)Mertes, Huber, Weitz, Heimerl, and
  Andr{\'e}]{mertesGANterfactualCounterfactualExplanations2022}
Silvan Mertes, Tobias Huber, Katharina Weitz, Alexander Heimerl, and Elisabeth
  Andr{\'e}.
\newblock {{GANterfactual}}---{{Counterfactual Explanations}} for {{Medical
  Non-experts Using Generative Adversarial Learning}}.
\newblock \emph{Frontiers in Artificial Intelligence}, 5, 2022.

\bibitem[Minh et~al.(2022)Minh, Wang, Li, and
  Nguyen]{minhExplainableArtificialIntelligence2022}
Dang Minh, H.~Xiang Wang, Y.~Fen Li, and Tan~N. Nguyen.
\newblock Explainable artificial intelligence: A comprehensive review.
\newblock \emph{Artificial Intelligence Review}, 55\penalty0 (5):\penalty0
  3503--3568, 2022.

\bibitem[Mothilal et~al.(2020)Mothilal, Sharma, and
  Tan]{mothilalExplainingMachineLearning2020}
Ramaravind~K. Mothilal, Amit Sharma, and Chenhao Tan.
\newblock Explaining machine learning classifiers through diverse
  counterfactual explanations.
\newblock In \emph{Proceedings of the 2020 {{Conference}} on {{Fairness}},
  {{Accountability}}, and {{Transparency}}}, pages 607--617, New York, NY, USA,
  2020. Association for Computing Machinery.

\bibitem[Nagesh et~al.(2023)Nagesh, Mishra, Naamad, Rehg, Shah, and
  Wagner]{nageshExplainingMachineLearning2023}
Supriya Nagesh, Nina Mishra, Yonatan Naamad, James~M. Rehg, Mehul~A. Shah, and
  Alexei Wagner.
\newblock Explaining a machine learning decision to physicians via
  counterfactuals.
\newblock In \emph{Proceedings of the {{Conference}} on {{Health}},
  {{Inference}}, and {{Learning}}}, pages 556--577. PMLR, 2023.

\bibitem[Narayanan et~al.(2018)Narayanan, Chen, He, Kim, Gershman, and
  {Doshi-Velez}]{narayananHowHumansUnderstand2018}
Menaka Narayanan, Emily Chen, Jeffrey He, Been Kim, Sam Gershman, and Finale
  {Doshi-Velez}.
\newblock How do {{Humans Understand Explanations}} from {{Machine Learning
  Systems}}? {{An Evaluation}} of the {{Human-Interpretability}} of
  {{Explanation}}, 2018.

\bibitem[Pegios et~al.(2024)Pegios, Feragen, Hansen, and
  Arvanitidis]{pegiosCounterfactualExplanationsRiemannian2024}
Paraskevas Pegios, Aasa Feragen, Andreas~Abildtrup Hansen, and Georgios
  Arvanitidis.
\newblock Counterfactual {{Explanations}} via {{Riemannian Latent Space
  Traversal}}, 2024.

\bibitem[Schutte et~al.(2021)Schutte, Moindrot, H{\'e}rent, Schiratti, and
  J{\'e}gou]{schutteUsingStyleGANVisual2021}
Kathryn Schutte, Olivier Moindrot, Paul H{\'e}rent, Jean-Baptiste Schiratti,
  and Simon J{\'e}gou.
\newblock Using {{StyleGAN}} for {{Visual Interpretability}} of {{Deep Learning
  Models}} on {{Medical Images}}, 2021.

\bibitem[Shrikumar et~al.(2017)Shrikumar, Greenside, and
  Kundaje]{shrikumarLearningImportantFeatures2017}
Avanti Shrikumar, Peyton Greenside, and Anshul Kundaje.
\newblock Learning {{Important Features Through Propagating Activation
  Differences}}.
\newblock In \emph{Proceedings of the 34th {{International Conference}} on
  {{Machine Learning}}}, pages 3145--3153. PMLR, 2017.

\bibitem[Song et~al.(2020)Song, Meng, and
  Ermon]{songDenoisingDiffusionImplicit2020}
Jiaming Song, Chenlin Meng, and Stefano Ermon.
\newblock Denoising {{Diffusion Implicit Models}}.
\newblock In \emph{International {{Conference}} on {{Learning
  Representations}}}, 2020.

\bibitem[Sturmfels et~al.(2020)Sturmfels, Lundberg, and
  Lee]{sturmfelsVisualizingImpactFeature2020}
Pascal Sturmfels, Scott Lundberg, and Su-In Lee.
\newblock Visualizing the {{Impact}} of {{Feature Attribution Baselines}}.
\newblock \emph{Distill}, 5\penalty0 (1):\penalty0 e22, 2020.

\bibitem[Sun et~al.(2025)Sun, Akman, and
  Schuller]{sunExplainableArtificialIntelligence2025}
Qiyang Sun, Alican Akman, and Bj{\"o}rn~W. Schuller.
\newblock Explainable {{Artificial Intelligence}} for {{Medical Applications}}:
  {{A Review}}.
\newblock \emph{ACM Transactions on Computing for Healthcare}, 6\penalty0
  (2):\penalty0 1--31, 2025.

\bibitem[Sundararajan et~al.(2017)Sundararajan, Taly, and
  Yan]{sundararajanAxiomaticAttributionDeep2017}
Mukund Sundararajan, Ankur Taly, and Qiqi Yan.
\newblock Axiomatic {{Attribution}} for {{Deep Networks}}.
\newblock In \emph{Proceedings of the 34th {{International Conference}} on
  {{Machine Learning}}}, pages 3319--3328. PMLR, 2017.

\bibitem[Tanyel et~al.(2025)Tanyel, Ayvaz, and
  Keserci]{tanyelKnownRealityExploiting2025}
Toygar Tanyel, Serkan Ayvaz, and Bilgin Keserci.
\newblock Beyond {{Known Reality}}: {{Exploiting Counterfactual Explanations}}
  for {{Medical Research}}, 2025.

\bibitem[{van der Velden} et~al.(2022){van der Velden}, Kuijf, Gilhuijs, and
  Viergever]{vanderveldenExplainableArtificialIntelligence2022}
Bas H.~M. {van der Velden}, Hugo~J. Kuijf, Kenneth G.~A. Gilhuijs, and Max~A.
  Viergever.
\newblock Explainable artificial intelligence ({{XAI}}) in deep learning-based
  medical image analysis.
\newblock \emph{Medical Image Analysis}, 79:\penalty0 102470, 2022.

\bibitem[Weng et~al.(2025)Weng, Pegios, Petersen, Feragen, and
  Bigdeli]{wengFastDiffusionBasedCounterfactuals2025}
Nina Weng, Paraskevas Pegios, Eike Petersen, Aasa Feragen, and Siavash Bigdeli.
\newblock Fast {{Diffusion-Based Counterfactuals}} for {{Shortcut Removal}} and
  {{Generation}}.
\newblock In \emph{European {{Conference}} on {{Computer Vision}}}, pages
  338--357, Cham, 2025. Springer Nature Switzerland.

\bibitem[Yeh et~al.(2019)Yeh, Hsieh, Suggala, Inouye, and
  Ravikumar]{yehInfidelitySensitivityExplanations2019}
Chih-Kuan Yeh, Cheng-Yu Hsieh, Arun Suggala, David~I Inouye, and Pradeep~K
  Ravikumar.
\newblock On the ({{In}})fidelity and {{Sensitivity}} of {{Explanations}}.
\newblock In \emph{Advances in {{Neural Information Processing Systems}}}.
  Curran Associates, Inc., 2019.

\bibitem[Zaher et~al.(2024)Zaher, Trzaskowski, Nguyen, and
  Roosta]{zaherManifoldIntegratedGradients2024}
Eslam Zaher, Maciej Trzaskowski, Quan Nguyen, and Fred Roosta.
\newblock Manifold integrated gradients: {{Riemannian}} geometry for feature
  attribution.
\newblock In \emph{Proceedings of the 41st {{International Conference}} on
  {{Machine Learning}}}, pages 58090--58104, Vienna, Austria, 2024. JMLR.org.

\bibitem[Zawacki et~al.(2019)Zawacki, Wu, Shih, Elliott, Fomitchev, Hussain,
  ParasLakhani, Culliton, and Bao]{siim-acr-pneumothorax-segmentation}
Anna Zawacki, Carol Wu, George Shih, Julia Elliott, Mikhail Fomitchev, Mohannad
  Hussain, ParasLakhani, Phil Culliton, and Shunxing Bao.
\newblock Siim-acr pneumothorax segmentation, 2019.
\newblock Kaggle.

\bibitem[Zhuo and Ge(2024)]{zhuoIG2IntegratedGradient2024a}
Yue Zhuo and Zhiqiang Ge.
\newblock {{IG2}}: {{Integrated Gradient}} on {{Iterative Gradient Path}} for
  {{Feature Attribution}}.
\newblock \emph{IEEE Transactions on Pattern Analysis and Machine
  Intelligence}, 46\penalty0 (11):\penalty0 7173--7190, 2024.

\end{thebibliography}
}

\clearpage

\setcounter{page}{1}
\maketitlesupplementary

\section{Theoretical analysis - Proofs}\label{sec:appendix_theory_proofs}

\paragraph{Preliminaries}

Let $\mathcal{C}:\mathbb{R}^d \to \mathbb{R}$ denote a continuously differentiable model (e.g., a class logit). 
For an input $x \in \mathbb{R}^d$ and a baseline $b \in \mathbb{R}^d$, the path-integrated gradient vector can be defined as
\begin{equation}
A(b)
:=
\int_0^1 
\nabla \mathcal{C}\big(b + \alpha (x-b)\big)
\, d\alpha.
\end{equation}
Then IG can be written compactly using the Hadamard (elementwise) product $\circ$:
\begin{equation}
\mathrm{IG}(x;b)
=
(x-b) \circ A(b).
\end{equation}

Integrated Gradients satisfies the \emph{completeness} property:
\begin{equation}
\sum_{i=1}^d \mathrm{IG}_i(x;b)
=
\mathcal{C}(x) - \mathcal{C}(b).
\end{equation}

\subsection{Perfect Counterfactual Baseline}\label{sec:appendix_prop1}

Let \(\mathcal{C} : \mathbb{R}^n \to \mathbb{R}\) be a differentiable function representing the scalar prediction of interest (e.g., the logit for a disease class). Let $x \in \mathbb{R}^n$ be a fixed input and let \(S \subseteq \{1,\dots,n\}\) denote the (unknown) set of \emph{pathological coordinates}, i.e., the coordinates where the input differs from a corresponding healthy, pathology-free version.

\begin{definition}{(Perfect Counterfactual Baseline)}{}
A baseline $\text{CF}^{*} \in \mathbb{R}^n$ is called a \emph{perfect counterfactual baseline} if $\mathcal{C}(\text{CF}^{*})$ corresponds to a neutral prediction, there exists a minimal causal index set $S \subseteq \{1,\dots,n\}$ such that \(x_i = \text{CF}^{*}_i \quad \text{for all } i \notin S;\), and the prediction difference \(\mathcal{C}(x) - \mathcal{C}(\text{CF}^{*})\) depends only on coordinates in $S$.
\end{definition}

Intuitively, $x^{cf}$ differs from $x$ only in those features causally responsible for the prediction. In this case, we can show that all attribution mass is concentrated on the causal coordinates $S$ only.

\begin{prop}[after skip=0pt]{(Perfect Attribution under Perfect Counterfactual Baseline)}{}
Let $\mathcal{C}$ be continuously differentiable and let $\text{CF}^{*}$ be a perfect counterfactual baseline for $x$. 
Then (1) $\mathrm{IG}_i(x;\text{CF}^{*}) = 0$ for all $i \notin S$ and (2) $\sum_{i \in S} \mathrm{IG}_i(x;\text{CF}^{*}) = \mathcal{C}(x) - \mathcal{C}(\text{CF}^{*})$.
\end{prop}
\begin{proof}{}{}
For any coordinate $i \notin S$, we have $x_i = x^{cf}_i$. 
By definition of IG,
\begin{equation}
\begin{aligned}
\mathrm{IG}_i(x;x^{cf})
=
(x_i - x^{cf}_i) \cdot \\
\int_0^1 
\frac{\partial \mathcal{C}(x^{cf} + \alpha(x-x^{cf}))}{\partial x_i}
\, d\alpha.
\end{aligned}
\end{equation}
Since $x_i - x^{cf}_i = 0$, it follows that
\begin{equation}
\mathrm{IG}_i(x;x^{cf}) = 0.
\end{equation}

By completeness,
\begin{equation}
\sum_{i=1}^d \mathrm{IG}_i(x;x^{cf})
=
\mathcal{C}(x) - \mathcal{C}(x^{cf}).
\end{equation}
Because only coordinates in $S$ differ between $x$ and $x^{cf}$, and all other coordinates receive zero attribution, we obtain
\begin{equation}
\sum_{i \in S} \mathrm{IG}_i(x;x^{cf})
=
\mathcal{C}(x) - \mathcal{C}(x^{cf}).
\end{equation}
Thus all attribution mass is concentrated on $S$.
\end{proof}

\subsection{Approximate counterfactuals}\label{sec:appendix_prop2}

In practice, a perfect counterfactual may not be available. 
We therefore consider approximate counterfactual baselines. For our theoretical analysis, we will make the following assumptions.

\begin{note}{(Assumptions)}{} We assume
\begin{enumerate}[leftmargin = 0.8cm]
    \item[(A1)] (\textbf{Lipschitz gradient}) 
    $\nabla \mathcal{C}$ is $L$-Lipschitz on the convex hull of $\{x, x^{cf}, b\}$:
    \[
    \|\nabla \mathcal{C}(u) - \nabla \mathcal{C}(v)\|_2 
    \le 
    L \|u - v\|_2.
    \]
    \item[(A2)] (\textbf{Bounded gradients}) 
    There exists $G_\infty < \infty$ such that along the integration paths
    \[
    \|\nabla \mathcal{C}(z)\|_\infty \le G_\infty.
    \]
\end{enumerate}
\end{note}

We show that, under these assumptions, if an approximate counterfactual baseline \(\tilde x\) is closer to the perfect counterfactual \(x^\star\) than a standard baseline \(b\) is, then the corresponding attribution \(\mathrm{IG}(x; \tilde x)\) is closer to the ideal attribution \(\mathrm{IG}(x; x^\star)\) in terms of error upper bound than \(\mathrm{IG}(x; b)\) is.

\begin{prop}[after skip=0pt]{(Closeness to True Counterfactual Implies Lower Attribution Error Upper Bound)}{}
Let $b \in \mathbb{R}^d$ be an arbitrary baseline. Define
\(\delta_b := \| b - \text{CF}^{*} \|_2\) and \(C := \frac{L}{2}\|x - \text{CF}^{*}\|_2 + G_\infty\). Under assumptions (A1) and (A2), for any baseline $b$,
\[
\|\mathrm{IG}(x;b) - \mathrm{IG}(x;\text{CF}^{*})\|_2
\le
C \, \delta_b.
\]
Consequently, for two baselines $b_1, b_2$, if
\(\|b_1 - \text{CF}^{*}\|_2 < \|b_2 - \text{CF}^{*}\|_2,\)
then their corresponding worst-case upper bounds satisfy
\(C \|b_1 - \text{CF}^{*}\|_2
<
C \|b_2 - \text{CF}^{*}\|_2\).
Hence, a baseline closer to the true counterfactual admits a strictly tighter theoretical upper bound on the attribution deviation.
\end{prop}

\begin{proof}
Using the compact form $\mathrm{IG}(x;b) = (x-b)\circ A(b)$, we write
\begin{equation}
\begin{split}
&\mathrm{IG}(x;b) - \mathrm{IG}(x;x^{cf}) \\
&\quad = (x-x^{cf}) \circ (A(b) - A(x^{cf})) \\
&\qquad + (x^{cf}-b) \circ A(b).
\end{split}
\end{equation}

\paragraph{Step 1: Bounding $\mathbf{A(b) - A(x^{cf})}$.}
For $\alpha \in [0,1]$, define
\begin{equation}
u(\alpha) = x^{cf} + \alpha(x-x^{cf}),
\quad
v(\alpha) = b + \alpha(x-b).
\end{equation}
Then
\begin{equation}
u(\alpha) - v(\alpha)
=
(1-\alpha)(x^{cf} - b),
\end{equation}
so
\begin{equation}
\|u(\alpha) - v(\alpha)\|_2
=
(1-\alpha)\delta_b.
\end{equation}

By the Lipschitz assumption,
\begin{equation}
\|\nabla \mathcal{C}(u(\alpha)) - \nabla \mathcal{C}(v(\alpha))\|_2
\le
L(1-\alpha)\delta_b.
\end{equation}

Integrating over $\alpha$,
\begin{equation}
\|A(b) - A(x^{cf})\|_2
\le
\int_0^1 L(1-\alpha)\delta_b \, d\alpha
=
\frac{L}{2}\delta_b.
\end{equation}

\paragraph{Step 2: Bounding the two terms.}

Using $\|u \circ v\|_2 \le \|u\|_2 \|v\|_\infty$:

\begin{align}
&\|(x-x^{cf}) \circ (A(b)-A(x^{cf}))\|_2 
\\
&\qquad \le
\|x-x^{cf}\|_2
\|A(b)-A(x^{cf})\|_\infty
\\
& \qquad \le
\frac{L}{2}\|x-x^{cf}\|_2 \delta_b.
\end{align}

Similarly,
\begin{align}
\|(x^{cf}-b) \circ A(b)\|_2
&\le
\|x^{cf}-b\|_2
\|A(b)\|_\infty
\\
&\le
\delta_b G_\infty.
\end{align}

\paragraph{Combining the bounds.}
\begin{align}
& \|\mathrm{IG}(x;b) - \mathrm{IG}(x;x^{cf})\|_2 \\
& \le
\left(
\frac{L}{2}\|x-x^{cf}\|_2 + G_\infty
\right)
\delta_b
=
C \delta_b.
\end{align}

The comparison between the bounds of two baselines follows immediately.
\end{proof}

\section{Empirical Validation of the Counterfactual Proximity Effect}\label{sec:appendix_proximity}

\begin{figure*}[h]
\centering
\includegraphics[width=\textwidth]{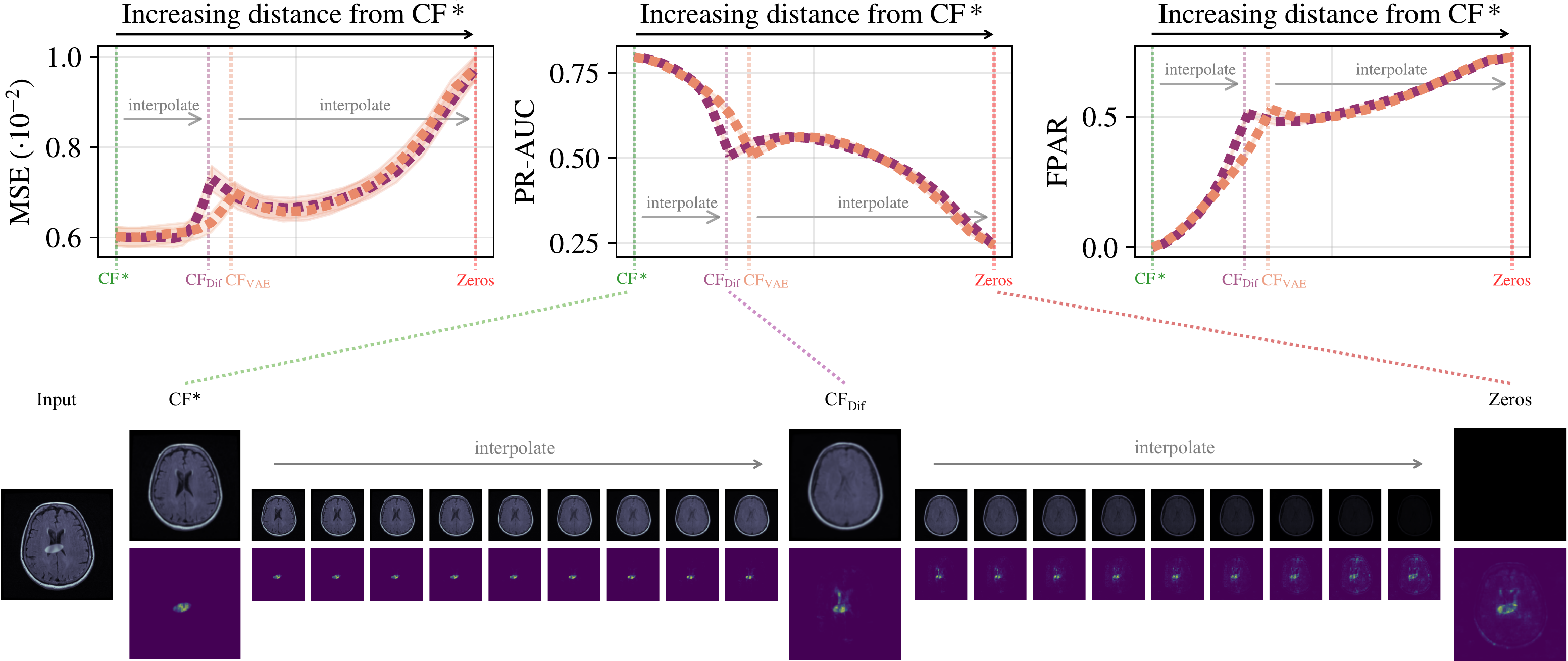}
\caption{
MSE, PR-AUC, and FPAR scores when increasing the distance to the perfect counterfactual $\mathrm{CF^*}$ by interpolating to the approximated counterfactual baselines and then interpolating to the zeros baseline.
}
\label{fig:distances_appendix}
\end{figure*}

To empirically assess how baseline distance from the true counterfactual affects attribution quality, we construct a controlled synthetic experiment based on the brain MRI dataset. Specifically, we take all healthy-class images and synthetically implant a tumor-shaped object into a randomly sampled half, yielding a binary classification task analogous to our other datasets. A classifier is then trained on this synthetic dataset following the same protocol.

The key advantage of this construction is that the unmodified healthy images serve as \emph{exact} counterfactuals $\text{CF}^*$, i.e.\ they are the true disease-free counterparts of the pathological inputs, with no generative approximation involved. This allows us to precisely control and measure the distance between a given baseline and the oracle counterfactual.

We then design a baseline interpolation trajectory that spans from $\text{CF}^*$ through our VAE- and diffusion-generated approximations and continues to the zero baseline, thereby constructing a continuum of baselines of increasing distance from the true counterfactual. Attribution quality metrics are computed at each point along this trajectory.

As shown in Fig.~\ref{fig:distances_appendix}, attribution quality degrades gradually as the baseline moves further from $\text{CF}^*$. This provides direct empirical support for our theoretical result: increasing the distance between a baseline and the true counterfactual leads to a measurable and systematic decrease in attribution quality. Taken together, the synthetic experiment confirms that the gains observed with our counterfactual baselines are not incidental, but follow from a principled relationship between baseline proximity and attribution faithfulness.

\section{Ablation tests}\label{sec:app_ablation}

As described in section \ref{sec_metrics_ablation}, an intuitive way to assess attribution quality are ablation tests, where certain pixels are replaced and the drop in classifier confidence is measured to assess their influence on the model output. In our experiments, we perform two common kinds of ablation tests \cite{sturmfelsVisualizingImpactFeature2020, ghorbaniInterpretationNeuralNetworks2019}: (1) Top-k ablation by removing the top-k most important pixels (provided by the different baselines) and measuring the drop in classifier confidence. (2) Region-based ablation by computing the attribution map’s center of mass and ablating increasingly large square regions around it, testing for coherent, clinically relevant localization.
We use two standard imputation methods to impute ablated pixels \cite{sturmfelsVisualizingImpactFeature2020}:
\begin{itemize}
    \item \textit{Mean Imputation}: replace each pixel with image-wide mean
    \item \textit{Midrange Imputation}: replace each pixel with the mean of the smallest and largest value in the image.
\end{itemize}
As these suffer from the same challenge when missingness is diagnostically meaningful, we additionally introduce the following more medically grounded imputation:
\begin{itemize}
    \item \textit{Mean Normal Imputation}: replace with pixelwise average from all healthy training samples.
\end{itemize}

We track how classifier confidence changes as more pixels are ablated. Faster confidence drops (i.e., lower curves) indicate more informative attributions.
However, the drop may stem from the imputations producing out-of-distribution inputs rather than from the removed pixels being important \cite{hookerBenchmarkInterpretabilityMethods2019}, and it is unclear which replacement values are truly neutral. Both issues are especially relevant in our setting, where the classifier detects subtle patterns in medical images that pixel replacements easily distort.
We therefore perform the ablation tests with an additional reference baseline. This reference is a mask-restricted attribution map by setting all \textsf{\small Zeros} attributions outside the ground-truth mask to zero, so that by construction it contains no false attributions. An ablation test that measures domain-specific faithfulness should yield low curves for this map.

\begin{figure*}[h]
\centering
\includegraphics[width=0.8\textwidth]{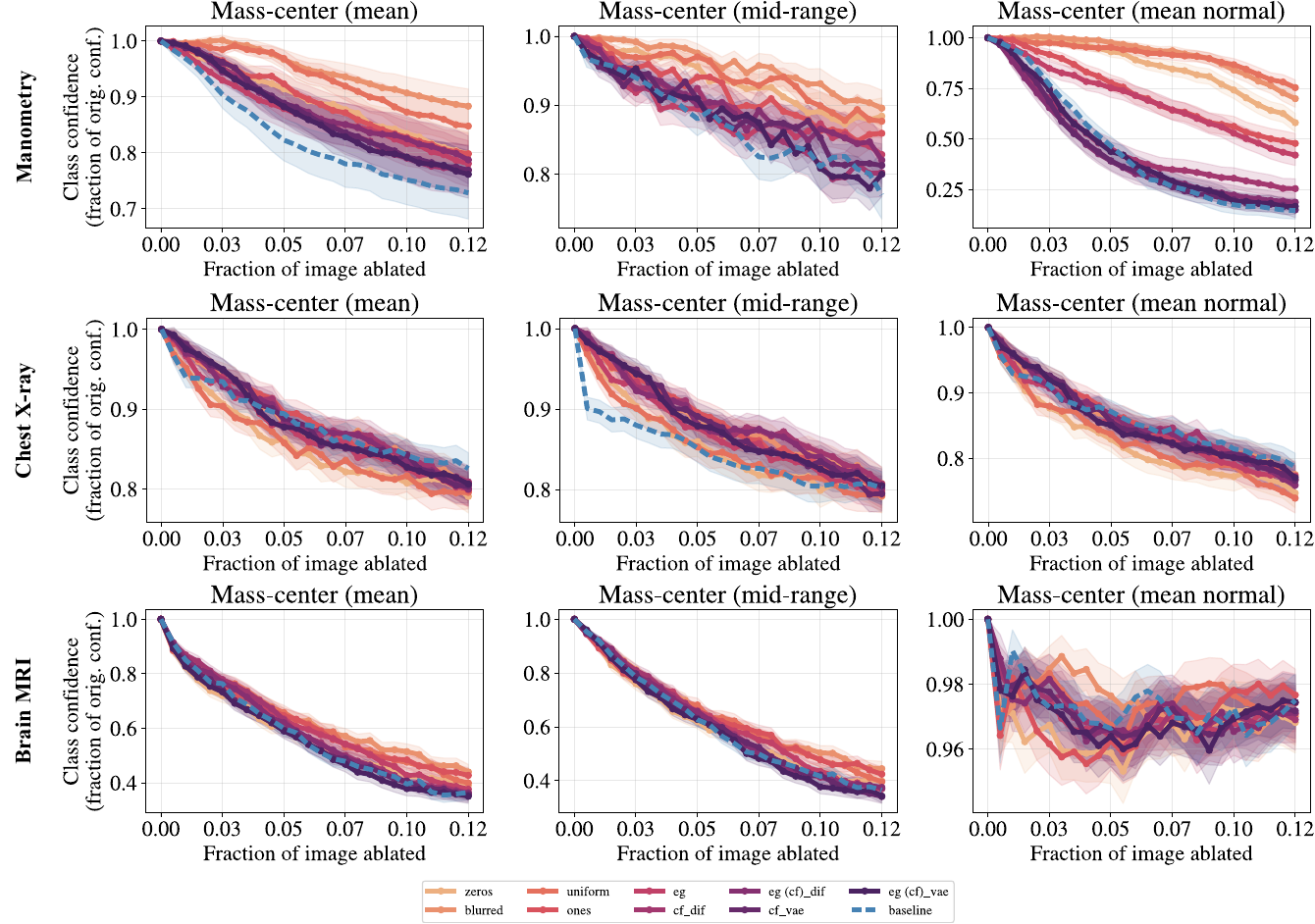}
\caption{
Results of the mass-center ablation tests
}
\label{fig:masscenter}
\end{figure*}

\begin{figure*}[h]
\centering
\includegraphics[width=0.8\textwidth]{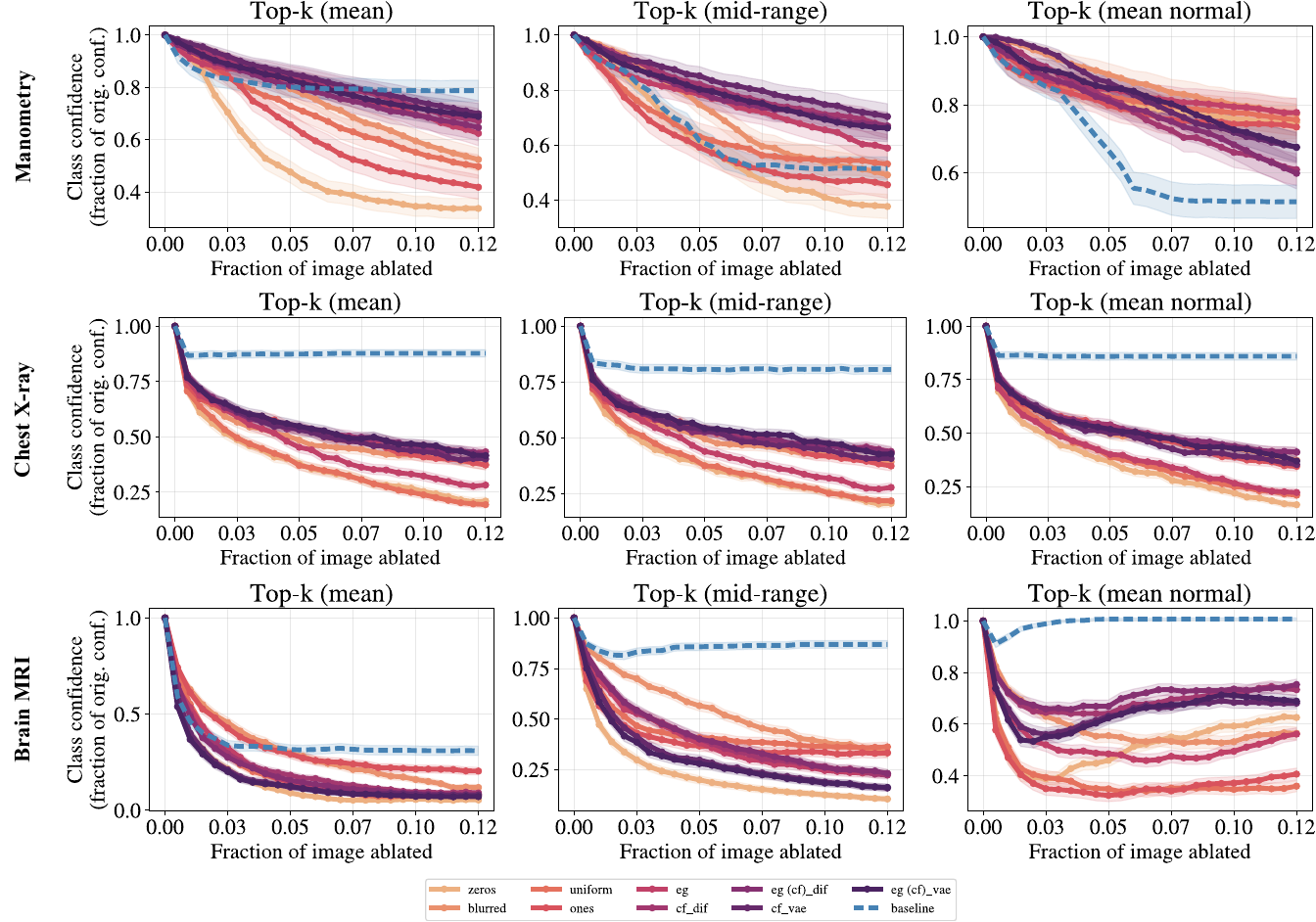}
\caption{
Results of the Top-k ablation tests
}
\label{fig:topk}
\end{figure*}

Fig.~\ref{fig:masscenter} shows the mass-center ablation results for the standard baselines and our proposed counterfactual baseline approaches, while Fig.~\ref{fig:topk} shows the top-k ablation results.

Two observations stand out. First, the results of the ablation tests vary with the imputation value used, indicating that the outcome of the test itself depends on this choice. However, there is no clear criterion for which imputation value yields reliable results, making it difficult to draw consistent conclusions from these tests. Second, the constructed reference baseline is not consistently among the lowest curves, particularly in the top-k ablation results. This suggests that the ablation tests may capture which pixels are important to the classifier, but not whether these pixels are medically meaningful, which is the focus of this paper. We therefore do not consider ablation tests a meaningful measure of attribution quality in our setting.

\section{Examples for qualitative evaluation}\label{appendix_sec_examples}
\begin{figure*}
\centering
\begin{subfigure}[t]{0.72\linewidth}
	\centering
	\includegraphics[width=\linewidth]{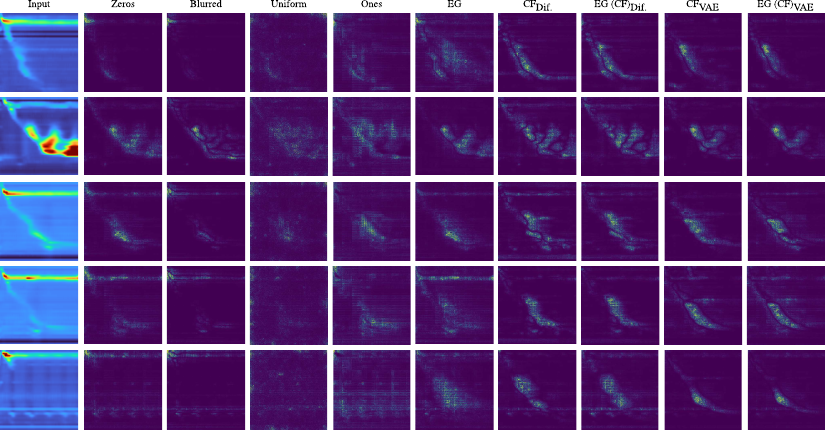}
        \caption{Manometry}
\end{subfigure}

\vspace{1em}

\begin{subfigure}[t]{0.72\linewidth}
	\centering
	\includegraphics[width=\linewidth]{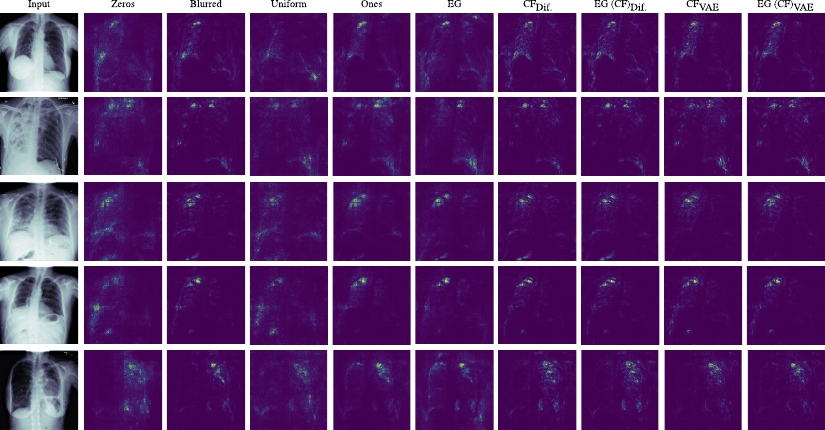}
        \caption{Chest X-ray}
\end{subfigure}

\vspace{1em}

\begin{subfigure}[t]{0.72\linewidth}
	\centering
	\includegraphics[width=\linewidth]{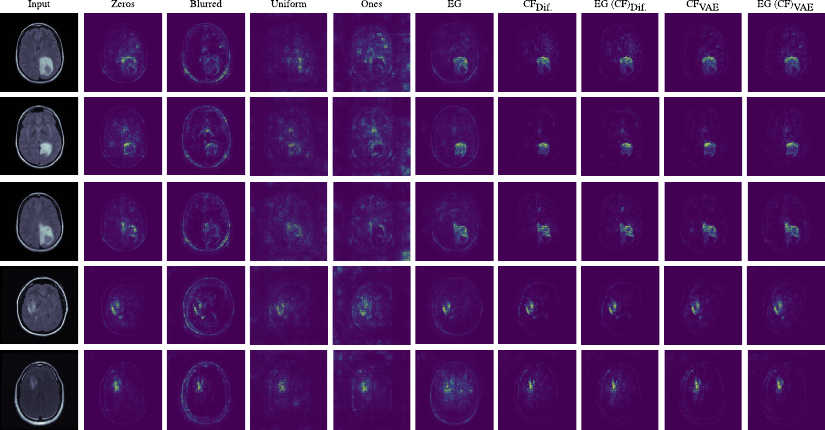}
        \caption{Brain MRI}
\end{subfigure}
\caption{Five examples from each data set for a qualitative comparison of the attributions obtained by the different baseline choices.}
\label{fig_appendix_examples}
\end{figure*}

Fig.~\ref{fig_appendix_examples} contains more examples from each data set for a qualitative evaluation and comparison of the attributions obtained by the different baselines.

\section{Comparison to pure counterfactual explanations - Using difference between input and counterfactual directly as the explanation}\label{appendix_difference}

As discussed in the main text, counterfactual explanation is a related concept in explainability research, where the counterfactual (or more specifically, the difference between the input and the counterfactual denoted as $(x - \hat{\text{CF}})$) directly serves as the explanation instead of using it as a baseline for another method like IG. One limitation is that a raw pixel-wise difference between the input and its counterfactual highlights regions of change, but it does not necessarily convey which of these changes are influential for the model’s prediction. This distinction is critical because not all observed differences are semantically meaningful or clinically relevant. For example, minor intensity variations or artifacts introduced during counterfactual generation may appear prominently in the difference map, even though they have negligible impact on the decision-making process. This can become problematic, as quality degrades rapidly when the counterfactual is imperfect. Generated counterfactuals are rarely perfect, and the raw difference then captures spurious variation across the entire image rather than localizing to the pathologically relevant region. Consequently, relying solely on raw differences risks misinterpretation by suggesting importance where none exists.

In contrast, path-based attribution methods such as IG and its extension EG provide a principled approach to quantifying feature importance. These methods compute attributions along a continuous interpolation path between a baseline and the input, effectively capturing how the model’s output changes as features transition from the baseline state to the observed state. This mechanism allows IG/EG to distinguish meaningful changes that influence the prediction from incidental variations. Furthermore, these methods satisfy key axiomatic properties, including sensitivity and implementation invariance \cite{sundararajanAxiomaticAttributionDeep2017}, which ensure theoretical soundness and consistency. A simple difference map does not guarantee these properties and therefore lacks the robustness required for reliable interpretability.

This advantage of IG based approaches becomes clear in the results, where the difference to the baseline (denoted \textsf{\small DIFF\textsubscript{Dif.}} and \textsf{\small DIFF\textsubscript{VAE}}) is resulting in significantly worse scores in almost any metric and data set. The overlap results can be seen in Table~\ref{tab:other_results}. Our baseline choices are significantly outperforming \textsf{\small DIFF\textsubscript{Dif.}} and \textsf{\small DIFF\textsubscript{VAE}} in almost any metric, with the only exception being PR-AUC on the manometry data set. 

\begin{figure*}
\centering
\begin{subfigure}[t]{0.50\linewidth}
	\centering
	\includegraphics[width=\linewidth]{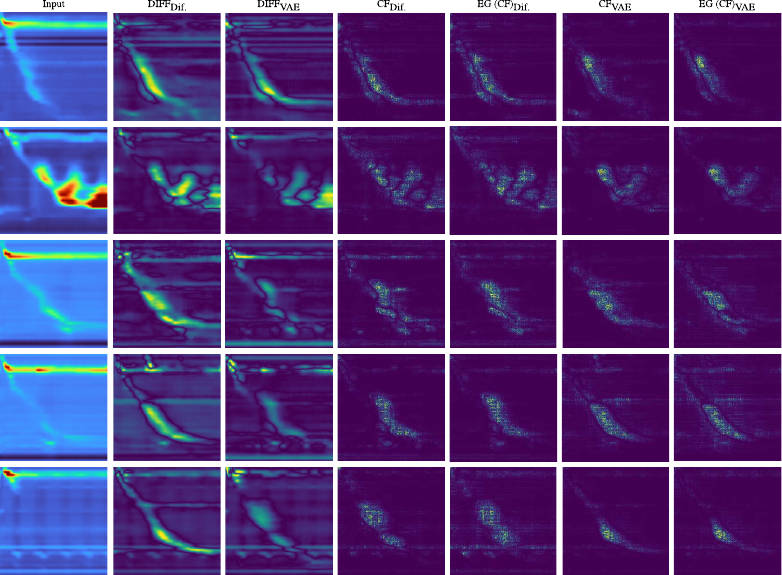}
        \caption{Manometry}
\end{subfigure}

\vspace{1em}

\begin{subfigure}[t]{0.50\linewidth}
	\centering
	\includegraphics[width=\linewidth]{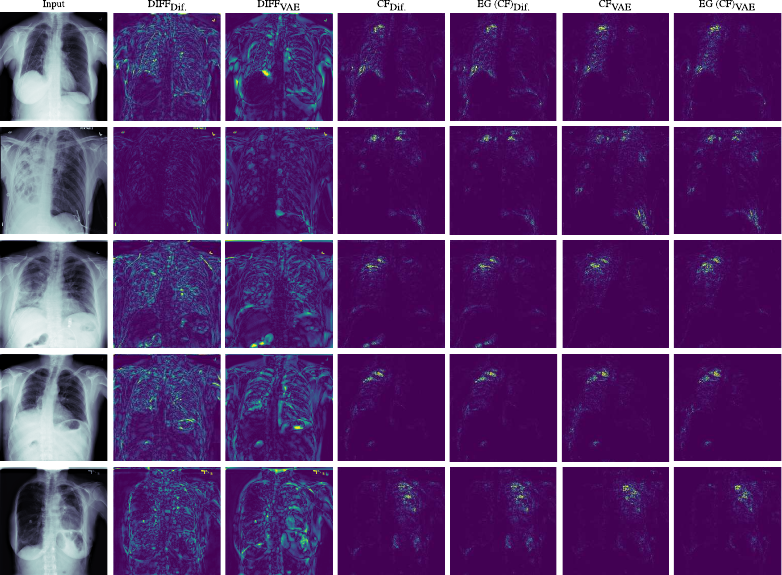}
        \caption{Chest X-ray}
\end{subfigure}

\vspace{1em}

\begin{subfigure}[t]{0.50\linewidth}
	\centering
	\includegraphics[width=\linewidth]{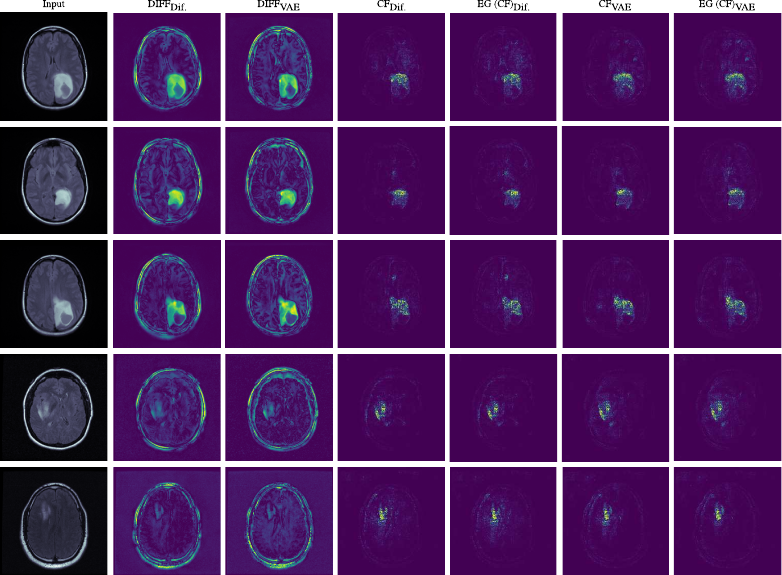}
        \caption{Brain MRI}
\end{subfigure}
\caption{Five examples from each data set for a qualitative comparison of direct counterfactual explanations (i.e. difference between counterfactual and input) and the attributions obtained by counterfactual baselines.}
\label{fig_appendix_examples_diff}
\end{figure*}

For a qualitative comparison, Fig.~\ref{fig_appendix_examples_diff} includes examples of using the difference map as attribution. Compared to the attributions generated by our proposed approaches, these maps highlight many regions beyond the region of interest, making it difficult to identify where the model actually focuses for its decision. This observation is consistent with our quantitative results.

\begin{figure*}
\centering
\begin{subfigure}[t]{0.8\linewidth}
	\centering
	\includegraphics[width=\linewidth]{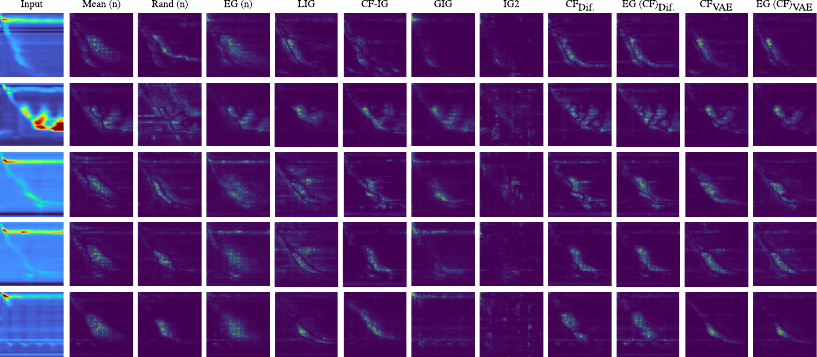}
        \caption{Manometry}
\end{subfigure}

\vspace{1em}

\begin{subfigure}[t]{0.80\linewidth}
	\centering
	\includegraphics[width=\linewidth]{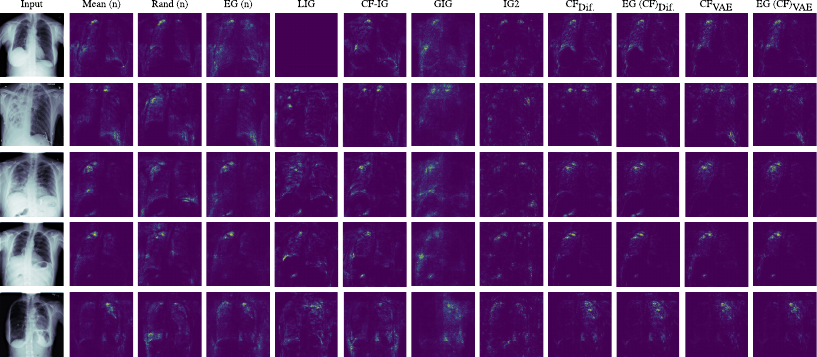}
        \caption{Chest X-ray}
\end{subfigure}

\vspace{1em}

\begin{subfigure}[t]{0.80\linewidth}
	\centering
	\includegraphics[width=\linewidth]{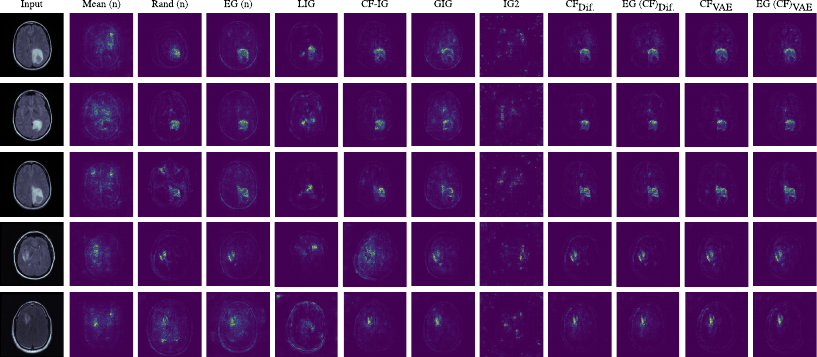}
        \caption{Brain MRI}
\end{subfigure}
\caption{Five examples from each data set for a qualitative comparison of the attributions obtained by different additional baseline choices and approaches.}
\label{fig_appendix_examples_additional}
\end{figure*}

\section{Comparison to other baseline choices, related works, and why a counterfactual is needed}\label{sec:appendix_otherstrategies}

\paragraph{Related approaches} As discussed in the main text, while our approach is conceptually distinct, yet grounded in similar underlying principles, we compare it to LIG~\cite{dravidMedXGANVisualExplanations2022} (see our implementation details below) and CF-IG~\cite{duellCounterfactualIntegratedGradientsCounterfactual2023}, as well as to the adaptive path methods GIG~\cite{kapishnikovGuidedIntegratedGradients2021}, and IG2~\cite{zhuoIG2IntegratedGradient2024a}. Table~\ref{tab:other_results} presents the results. As can be seen across all evaluated datasets, our baselines consistently outperform the other approaches in terms of attribution quality. For a qualitative comparison, Fig.~\ref{fig_appendix_examples_additional} includes examples of using the related concepts. When compared to the attributions generated by our proposed approaches, more spurious and less precise attributions can be observed.

\paragraph{Other alternative baselines} Additionally, as highlighted in the main text, one might consider using samples from the normal class as a more natural and computationally efficient baseline for attribution methods such as IG. This approach would avoid the need to generate a dedicated counterfactual for each input and instead rely on representative samples from the distribution of healthy or non-anomalous data. To evaluate the effectiveness of this strategy, we conduct additional experiments using three alternative baseline choices:
\begin{itemize}
    \item \textsf{\small Mean (n)}: the mean of all normal samples
    \item \textsf{\small Rand (n)}: a single random normal sample
    \item \textsf{\small EG (n)}: EG with 50 random normal samples
\end{itemize}

The results of these experiments, also presented in Table \ref{tab:other_results}, reveal that these baselines are consistently outperformed by the counterfactual-based approaches \textsf{\small CF} and \textsf{\small EG (CF)}. This performance gap highlights the limitations of using generic normal samples, which may not be sufficiently tailored to the specific input being explained. In particular, normal samples may differ significantly in structure, intensity, or semantic content from the input, leading to attribution maps that are less focused and less informative.
For a qualitative comparison, Fig.~\ref{fig_appendix_examples_additional} includes examples of using the alternative approaches. When compared to the attributions generated by our proposed approaches, more spurious and less precise attributions can be observed. These results further strengthen our argument that a baseline tailored to the input is achieving more accurate attribution results.

\paragraph{Implementation of Latent Integrated Gradients (LIG)}\label{sec:appendix-lig}
We provide a short explanation of LIG as well as our implementation details of it. The original definition of LIG by \citet{dravidMedXGANVisualExplanations2022} is formulated as:

\begin{equation}
\begin{aligned}
\mathrm{LIG} = \left[\mathcal{G}(z_1, z_2) - \mathcal{G}(z_1, \mathbf{0})\right] \cdot \\ \int_{\alpha=0}^1 \frac{\partial \mathcal{C}\left(\mathcal{G}(z_1, \mathbf{0} + \alpha z_2)\right)}{\partial x} \, d\alpha
\end{aligned}
\end{equation}
$\mathcal{G}$ is the generator of a GAN that takes two input latent vectors $z_1, z_2$, with $z_1$ corresponding to anatomical structure and $z_2$ corresponding to pathology features. Consequently, $\mathcal{G}(z_1, z_2)$ represents a ``positive'' (pathological) realization, while $\mathcal{G}(z_1, \mathbf{0})$ represents a ``negative'' (counterfactual healthy) realization. The LIG is then obtained by integrating gradients along an interpolation path in the latent space between these two realizations.
A key limitation of this formulation is that the interpolation path consists entirely of synthetic samples, none of which correspond to the original input image. As a result, the quality and reliability of the attributions depend heavily on the generator’s ability to produce realistic and semantically consistent outputs. This assumption is particularly daring in the medical imaging domain, where even subtle deviations from real data can distort pathological cues and lead to misleading interpretations.

For our comparison, we adjusted the formulation of LIG to work with our VAE setup instead of the original GAN setup. Specifically, we compute LIG with:
\begin{equation}
\begin{aligned}
\mathrm{LIG} = \left[\mathcal{D}(z_0) - \mathcal{D}(z^*)\right] \cdot \\ \int_{\alpha=0}^1 \frac{\partial \mathcal{C}\left(\mathcal{D}(z^*) + \alpha \cdot (\mathcal{D}(z_0) -  \mathcal{D}(z^*))\right)}{\partial x} \, d\alpha
\end{aligned}
\end{equation}
where $z_0$ is the latent encoding of input $x$, i.e. $z_0 = \mathcal{E}(x)$, and $z^*$ is the latent representation of the counterfactual.

We acknowledge that LIG’s performance is highly dependent on the quality of generated realizations, where a stronger generative model could improve attribution quality. However, this reliance underscores a fundamental limitation of LIG: the original input is excluded from the attribution process, making the explanation entirely dependent on the synthetic samples.

\section{Limitations}\label{sec:limitations}
While our approach improves attribution faithfulness in most evaluated scenarios, it introduces certain limitations. Computing baselines is more complex than using predefined ones, as our method requires training and sampling from a generative model, increasing computational overhead. We argue this trade-off is justified in medical contexts. While training the VAE and diffusion model incurs an upfront cost (dependent on hardware and training details), this is amortized over inference, where generating a counterfactual for a given input is lightweight. To demonstrate the computational efficiency, we measure training times of the VAE and diffusion model and inference times for computing counterfactual baselines. The results are reported in appendix~\ref{sec:appendix_computational_cost}.

Additionally, the quality of generated baselines depends on the performance of the generative model. When the model fails to produce realistic or semantically meaningful counterfactuals, as observed especially in the Chest X-ray dataset, the baselines may not accurately represent the absence of class-defining features, limiting attribution reliability. In this aspect, we acknowledge that counterfactuals are artificial constructs and do not claim them as ground truth, but rather as more informative and plausible references than conventional baselines.
Lastly, the interpretability of any explanation is constrained by the quality of the underlying classifier. While our classifiers performed reasonably well (see appendix~\ref{sec:appendix_classifier}), even well-calibrated attributions are only as reliable as the model they explain.

\section{Quantitative assessment of counterfactual realism and its impact to attribution quality}\label{sec:appendix_cf_quality}
\label{sec:cf_realism}

Our theoretical framework (Proposition~\ref{prop:approx-counterfactual-better}) establishes that attribution quality improves as the generated counterfactual approaches the true counterfactual. Since the true counterfactual is unobservable for real medical data, we instead assess two properties to approximate the quality of our generated counterfactuals: whether the counterfactuals are \emph{distributionally aligned} with the real healthy manifold, i.e., closer to it than the input, and whether the changes are \emph{spatially targeted} at the diagnostically relevant regions. We find empirical evidence for both, supporting the operational realization of semantic missingness. This indicates that our generators produce genuine counterfactuals, and that the attribution improvements therefore stem from counterfactual changes rather than from arbitrary samples that merely happen to yield better attributions.

\begin{figure*}
\centering
\includegraphics[width=0.9\linewidth]{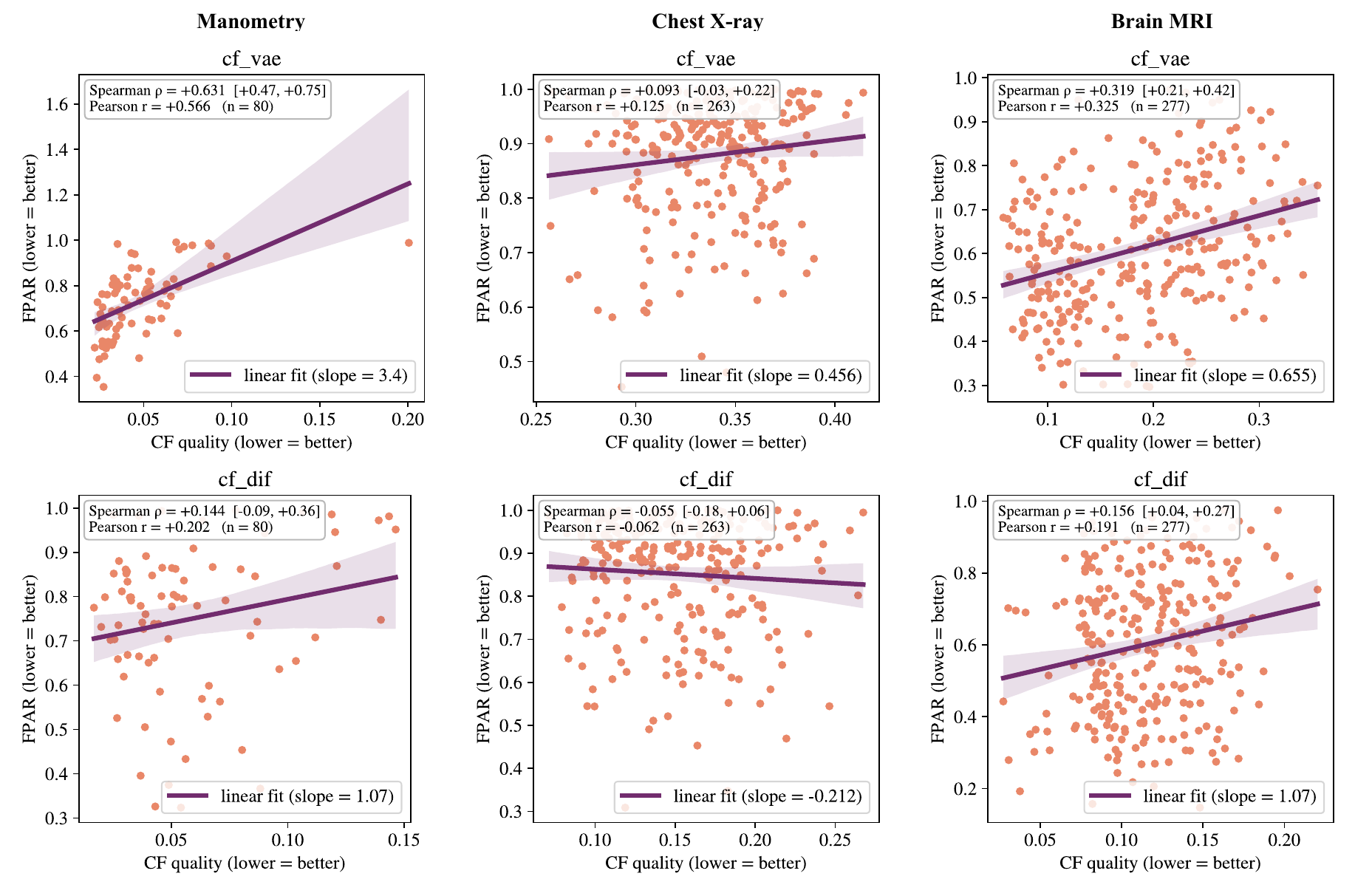}
\caption{
Correlations of approximated counterfactual quality to FPAR metric. A better FPAR metric is generally associated with a lower higher approximated counterfactual quality.
}
\label{fig:appendix_correlation_fpar_quality}
\end{figure*}

\paragraph{Distributional realism via classifier-feature FID.}
We measure the distance between the distribution of generated counterfactuals and the distribution of real healthy images, using a modified Fréchet Inception Distance (FID) by  employing the penultimate-layer features of the trained classifier (instead of a separate Inception model) as the embedding space. Working in the classifier's feature space ensures that distances reflect variation along axes the model itself uses to discriminate healthy from pathological inputs, which is the relevant notion of similarity for our setting.

To anchor the resulting numbers, we compute three reference distances: (i) FID between two disjoint splits of the real healthy set, which estimates the within-distribution noise floor; (ii) FID between real healthy and real pathological images, which provides an upper anchor reflecting the natural separation between the two classes; and (iii) FID between generated counterfactuals and the real healthy set, which measures how far our generated samples sit from the healthy distribution. Results for the three dataset are reported in Table~\ref{tab:fid}.

\begin{table}
\scriptsize
\setlength{\tabcolsep}{4pt}
\caption{Classifier-feature FID between distributions across the three medical datasets. Lower values indicate greater distributional similarity. Both generators reduce the distance to the healthy distribution relative to the pathological inputs.}
\label{tab:fid}
\centering
\begin{tabular}{L{3.0cm}C{1.2cm}C{1.3cm}C{1.2cm}}
\toprule
\textbf{Comparison} & \textbf{Manometry} & \textbf{Chest X-ray} & \textbf{Brain MRI} \\
\midrule
Real healthy vs.\ real healthy (noise floor) & 51.5 & 20.7 & 15.8 \\
\hdashline[1pt/1pt]
\noalign{\vskip 0.3ex}
Real healthy vs.\ real pathological (upper anchor) & 1038.2 & 414.1 & 1239.8 \\
\hdashline[1pt/1pt]
\noalign{\vskip 0.3ex}
Real healthy vs.\ diffusion counterfactuals & 599.1 & 348.0 & 813.7 \\
\hdashline[1pt/1pt]
\noalign{\vskip 0.3ex}
Real healthy vs.\ VAE counterfactuals & 443.2 & 388.1 & 291.1 \\
\bottomrule
\end{tabular}
\end{table}

Both generators reduce the distributional distance to the real healthy set relative to the pathological inputs on all data sets, indicating that the generated counterfactuals meaningfully shift the input toward the healthy manifold rather than producing arbitrary or off-manifold artifacts. The residual gap to the noise floor reflects a combination of finite-sample bias inherent to FID estimation, anatomy anchoring (counterfactuals are conditioned on pathological inputs and therefore retain patient-specific anatomical structure), and the natural imperfection of the generative models. Importantly, our framework does not require this gap to vanish: Proposition~\ref{prop:approx-counterfactual-better} guarantees that attribution quality improves continuously as the counterfactual approaches the true reference, and the framework is generator-agnostic, so improvements in counterfactual generation translate directly into improvements in attribution quality.

\begin{figure*}
\centering
\includegraphics[width=0.8\linewidth]{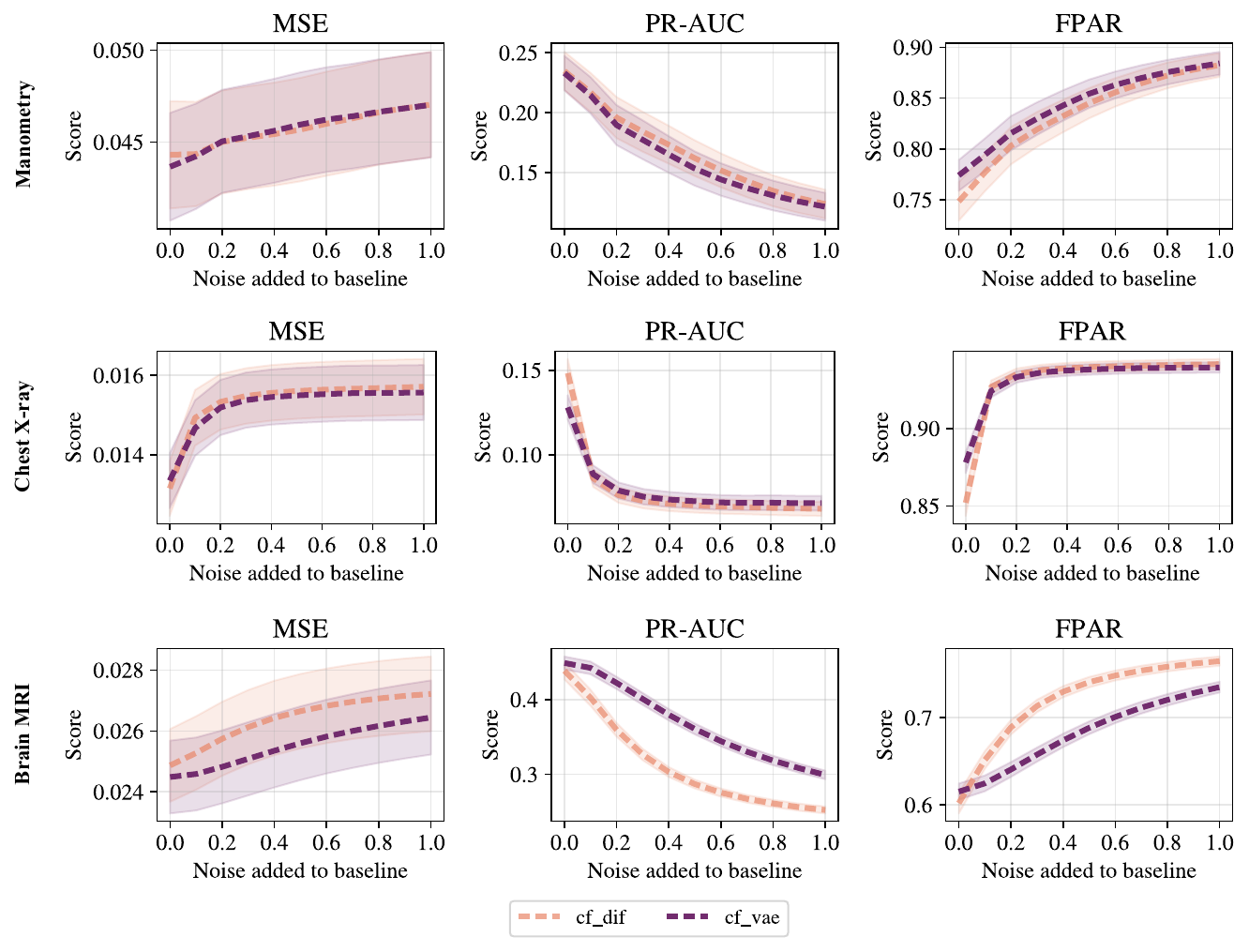}
\caption{
MSE, PR-AUC, and FPAR when adding noise to the generated counterfactuals, thus degrading the counterfactual quality. The attribution quality metrics get gradually worse with increased noise.
}
\label{fig:appendix_noise_degradation}
\end{figure*}

\paragraph{Spatial localization of counterfactual changes.}
A second, complementary criterion for a meaningful counterfactual is that the modifications introduced by the generator are concentrated in the diagnostically relevant region rather than distributed globally across the image. To assess this, we compute the average per-pixel change between the input $x$ and the generated counterfactual $x_\text{normal}$, separately inside and outside the ground-truth pathology mask, and report the ratio:
\[
r = \frac{\frac{1}{|M|} \sum_{i \in M} |x_i - x_{\text{normal},i}|}{\frac{1}{|\bar M|} \sum_{i \in \bar M} |x_i - x_{\text{normal},i}|},
\]
where $M$ denotes the pathology mask and $\bar M$ its complement. A ratio $r > 1$ indicates that changes are concentrated inside the clinically relevant region. Across all three datasets and both generators, we observe consistently $r > 1$ (Table~\ref{tab:localization}), confirming that the generators selectively modify the pathological region while leaving surrounding anatomy largely intact.

\begin{table}
\scriptsize
\setlength{\tabcolsep}{4pt}
\caption{Ratio of average per-pixel change inside vs.\ outside the ground-truth pathology mask (mean $\pm$ se). Values $>1$ indicate that the generators concentrate modifications in clinically relevant regions.}
\label{tab:localization}
\centering
\begin{tabular}{L{3cm}C{1.3cm}C{1.3cm}C{1.3cm}}
\toprule
\textbf{Comparison} & \textbf{Manometry} & \textbf{Chest X-ray} & \textbf{Brain MRI} \\
\midrule
Diffusion counterfactuals & 3.15 $\pm$ 0.35 & 2.07 $\pm$ 0.08 & 5.37 $\pm$ 0.35 \\
VAE counterfactuals & 3.21 $\pm$ 0.18 & 1.70 $\pm$ 0.03 & 3.55 $\pm$ 0.10 \\
\bottomrule
\end{tabular}
\end{table}

\paragraph{Impact of counterfactual realism to attribution quality}
As a more direct evaluation of Proposition~\ref{prop:approx-counterfactual-better}, we also assess the impact of counterfactual realism to attribution quality by measuring the correlation between a per-sample counterfactual realism score and the corresponding attribution quality metric FPAR. However, as there is no true ‘ground-truth’ counterfactual available for our three datasets, the quality metric and the depending results need to be viewed carefully. We approximate the counterfactual quality as the mean cosine distance to the k-nearest (in our current experiments, k = 5), representations of healthy samples in feature space of the classifier. The correlations can be seen in Fig.~\ref{fig:appendix_correlation_fpar_quality}, together with a scatter plot showing the approximated counterfactual quality vs. FPAR.

As observed, all correlations are positive except for the diffusion-based counterfactual approach on the chest data set, indicating that a better counterfactual indeed generally results in a better attribution quality. 
However, as we don’t know the ground truth counterfactual, this test relies on an approximation of counterfactual quality. 
Therefore, to further strengthen our claim, we additionally perform a test similar to the synthetic proximity test in section \ref{sec:appendix_proximity}, trying to decrease the quality of our generated counterfactual by gradually adding random gaussian noise, where the last step contains the amount of noise such that the distance to the original counterfactual is similar to the distance of the zeros baseline to the original counterfactual. This aims to simulate a gradual decrease in counterfactual quality. Per noise step we present the mean (± se) for the metrics MSE, PR-AUC and FPAR, giving a depiction by how much the attribution quality decreases when the counterfactual quality degrades. The results are shown in Fig.~\ref{fig:appendix_noise_degradation}, demonstrating that with increased noise, all attribution quality metrics get gradually worse on every data set, further supporting the theoretical findings of our work.

\paragraph{Summary.}
Together, the analyses show that the generated counterfactuals lie  closer to the real healthy distribution than the pathological inputs and differ from the input primarily in the diagnostically relevant region. Additionally, they demonstrate that a higher counterfactual quality is generally associated with a higher attribution quality. This provides empirical support that the generators produce genuine counterfactuals which are operationally consistent with the notion of semantic missingness introduced in our work, while also supporting our theoretical claims on the real data sets.

\section{Dynamic counterfactual baseline generation}\label{sec:appendix_counterfactual_generation}

\subsection{Dynamic counterfactual baseline generation using VAEs}\label{sec_method_cf_vae}
\begin{figure}
\centering
\includegraphics[width=\linewidth]{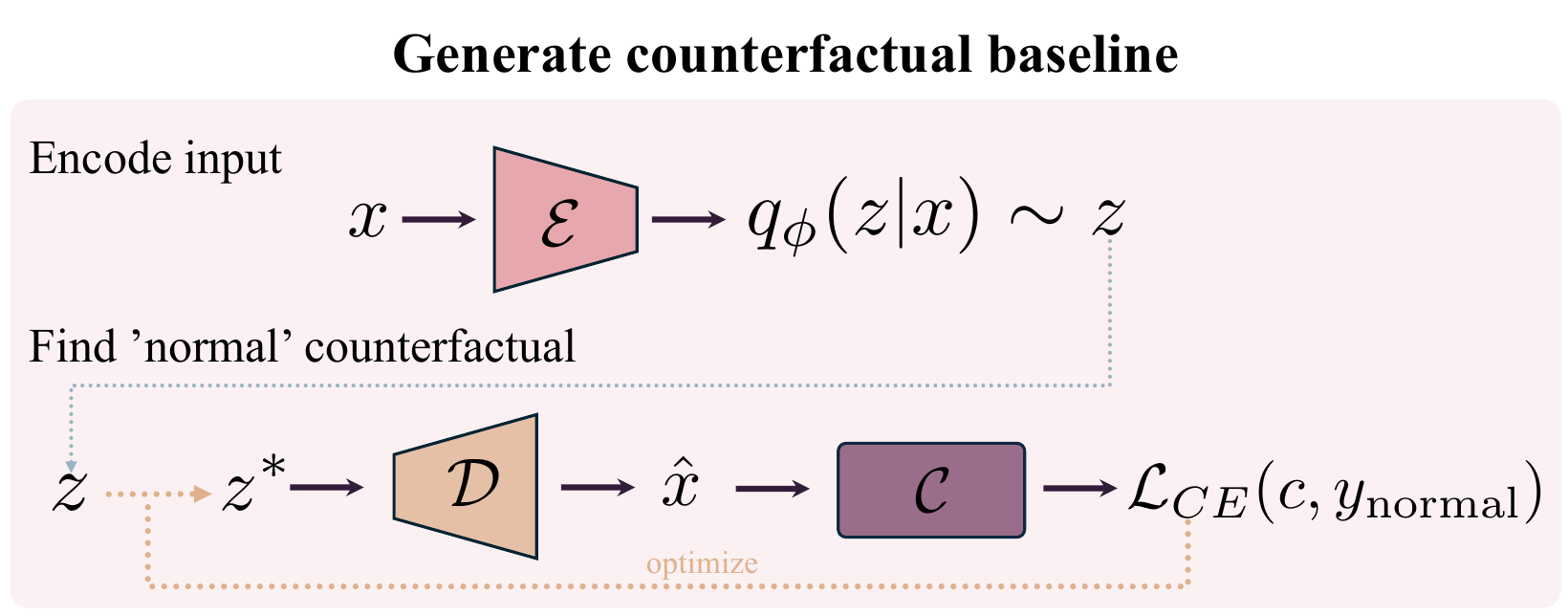}
\caption{
Overview of our approach to generate a counterfactual baseline $z^*$ for input sample $x$.
}
\label{fig:setup}
\end{figure}

This section contains more details on the optimization in the latent space of the VAE to generate the counterfactual baseline, as well as examples of the optimization path from pathological to counterfactual sample.

For a given input image $x \in \mathbb{R}^n$, we aim to find a ``normal'' counterfactual $\hat{x}^*$. To this end, we use a simple Convolutional Neural Network (CNN)-based VAE \cite{kingmaAutoEncodingVariationalBayes2014}. The VAE consists of an encoder $\mathcal{E}: \mathbb{R}^n \rightarrow \mathbb{R}^d$ and decoder $\mathcal{D}: \mathbb{R}^d \rightarrow \mathbb{R}^n$ and is trained to reconstruct input samples, i.e. $\mathcal{D}(\mathcal{E} (x)) = \hat{x} \approx x$. To obtain a latent representation $z \in \mathbb{R}^d$ of an input $x$, we can sample $z \sim q_\phi(z \mid x) \coloneqq \mathcal{N}(\mu_\phi(x), \sigma^2_\phi(x) I), \text{with} (\mu_\phi(x), \sigma^2_\phi(x)) = \mathcal{E}(x)
$. The architecture details of the VAE can be found in appendix \ref{sec:appendix_vae}.

The classifier $\mathcal{C}: \mathbb{R}^n \rightarrow \mathbb{R}^c$ maps the input to the predicted output logits for each possible output class, with one of the classes denoting the ``healthy'' (i.e. normal) class. In our experiments, we use a SwinTransformer-based \cite{liuSwinTransformerV22022} model for the classifier. The detailed architecture of the classifier can be found in appendix \ref{sec:appendix_classifier}.

To obtain a suitable counterfactual baseline $\hat{x}^*$ for input $x$, we aim to find the latent code $z^*(x)$ that, when decoded, minimizes the cross entropy loss to the normal class target $y_n$ but is still sufficiently close to the input $x$ otherwise, which can be formulated as:
\begin{equation}
\label{eq:argmin}
\begin{aligned}
z^*(x) \; = \; \arg\min_{z \in \mathbb{R}^d} \biggl[ \mathcal{L}_{CE}(\mathcal{C}(\mathcal{D}(z)),y_n) + \quad \\ \lambda_{\text{sim}} \: \| x - \mathcal{D}(z)) \|_2^2 \biggr]
\end{aligned}
\end{equation}
We perform this optimization by first obtaining an initial latent representation $z_0$ of the input $x$ by $z_0 = \mathcal{E}(x)$.
Then, we use gradient descent on $z$ to minimize the cross entropy loss function $\mathcal{L}_{CE}(\mathcal{C}(\mathcal{D}(z_i)),y_{\text{normal}})$ using the Adam optimizer. This optimization proceeds until a confidence threshold of the normal class is reached:
\begin{equation}
z_{t+1} = z_t - \eta_t \cdot \mathtt{Adam} \Bigl[ \nabla_z \mathcal{L}_{CE} \bigl(\mathcal{C}(\mathcal{D}(z)),y_n \bigr) \Bigr]
\end{equation}
In our implementation, we omit an explicit similarity term as in equation \ref{eq:argmin} by setting $\lambda_{\text{sim}} = 0$, since we are starting at $z_0 = \mathcal{E}(x)$ (where $\mathcal{D}(z_0))$ is already very close to $x$). During optimization, we stop once the desired confidence threshold is reached.
The decoded image $\hat{x}^* = \mathcal{D}(z^*)$ is then used as the counterfactual baseline. An overview of the whole process can also be seen in Fig.\ref{fig:setup}.

\subsubsection{Optimization details}

As described in section \ref{sec_method_cf_vae}, the counterfactual is found by traversing the latent space of the VAE (using the Adam optimizer) aiming to minimize the cross entropy loss between the decoded sample's predicted class and the true label. Table \ref{tab_optimization_hyperparameters} shows the hyperparameters used to find the counterfactual. The parameters were chosen based on multiple runs and deemed suitable enough for our experiments.

\begin{table}
\small
\caption{Optimization parameters}
\label{tab_optimization_hyperparameters}
\centering
\begin{tabular}{@{\hskip 0.1em}l @{\hskip 0.2em} c @{\hskip 0.8em} c @{\hskip 0.8em} c  @{\hskip 0.2em}}
\toprule
 & \multicolumn{3}{c}{\textbf{Datasets}} \\
\cmidrule(r){2-4} 
\textbf{Parameter} & \textbf{Manometry} & \textbf{Chest X-ray} & \textbf{Brain MRI} \\
\midrule
Learning rate & 0.1 & 0.02 & 0.1 \\
Max. iterations & 50 & 50 & 50 \\
Confidence threshold & 0.99 & 0.99 & 0.99 \\
\bottomrule
\end{tabular}
\end{table}

\subsubsection{Examples of path from pathological sample to normal counterfactual sample}
\begin{figure*}
\centering
\begin{subfigure}[t]{\linewidth}
	\centering
	\includegraphics[height=1.3cm]{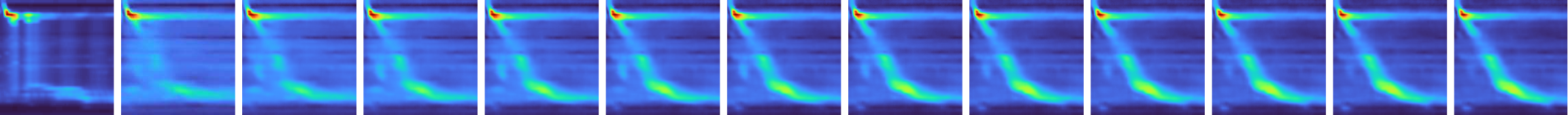}
        \caption{Manometry}
\end{subfigure}

\vspace{1em}

\begin{subfigure}[t]{\linewidth}
	\centering
	\includegraphics[height=1.3cm]{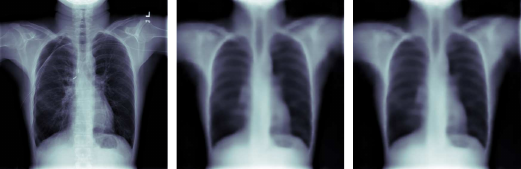}
        \caption{Chest X-ray}
\end{subfigure}

\vspace{1em}

\begin{subfigure}[t]{\linewidth}
	\centering
	\includegraphics[height=1.3cm]{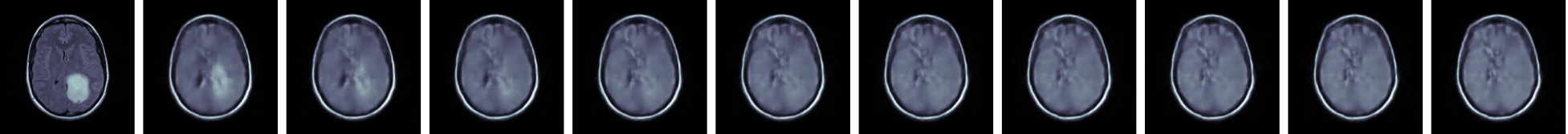}
        \caption{Brain MRI}
\end{subfigure}
\caption{Examples of decoded samples along the optimization path from the pathological sample to the counterfactual sample.}
\label{fig_examples_path}
\end{figure*}

For illustration purposes, we show examples of decoded samples along the optimization path in Fig.~\ref{fig_examples_path}. The final counterfactual is reached once the specified confidence threshold is reached.

\subsection{Dynamic counterfactual baseline generation using diffusion models}\label{sec_method_cf_diff}
\begin{figure}
\vspace{-1em}
\centering
\includegraphics[width=\linewidth]{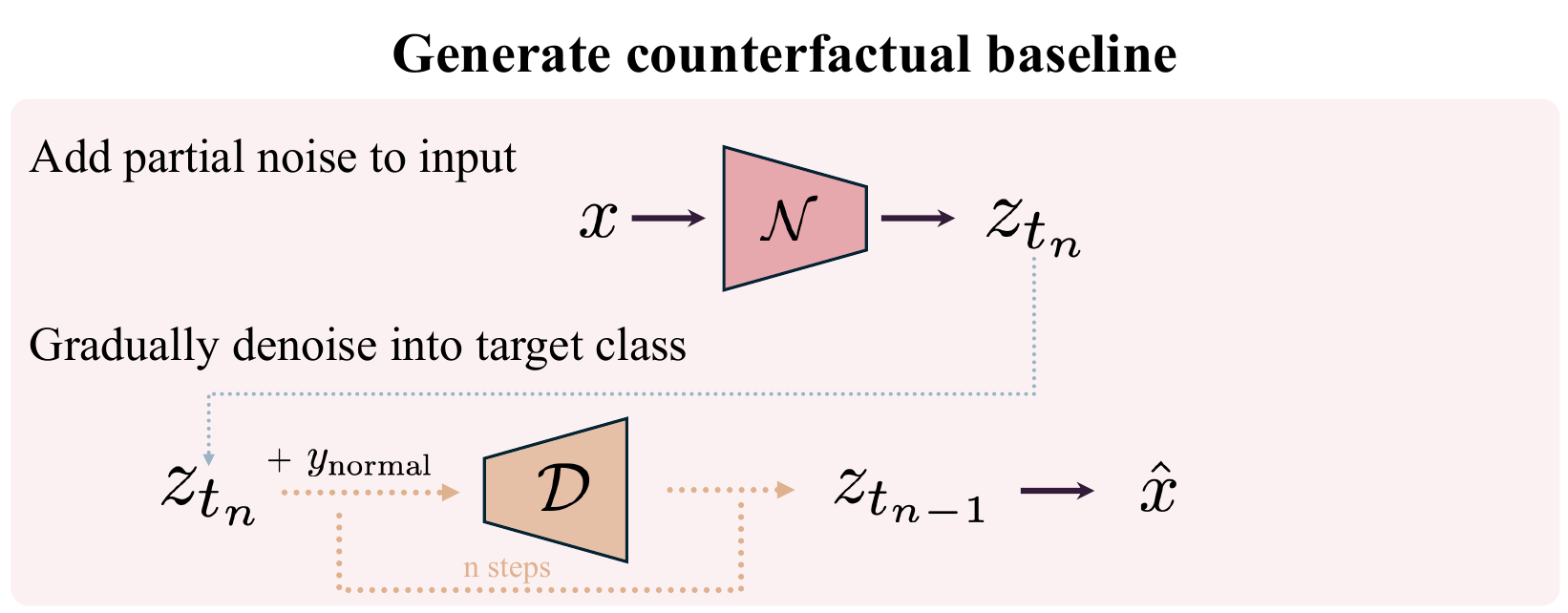}
\caption{
Overview of our approach to generate a counterfactual baseline $z^*$ for input sample $x$.
}
\label{fig:setup_diff}
\end{figure}
For a given input image $x \in \mathbb{R}^n$, we aim to find a ``normal'' counterfactual $\hat{x}^*$ using a diffusion model operating directly in pixel space. The approach is based on a conditional denoising diffusion probabilistic model (DDPM) \cite{hoDenoisingDiffusionProbabilistic2020}, in which a UNet-based denoising network $\epsilon_\theta: \mathbb{R}^n \times \mathbb{R} \times \mathcal{Y} \rightarrow \mathbb{R}^n$ is trained to predict the noise $\epsilon$ added to an input $x$ at each diffusion timestep $t$, conditioned on a class label $y \in \mathcal{Y}$, by minimising:

\begin{equation}
    \mathcal{L}_{\text{diff}} = \mathbb{E}_{x, \epsilon \sim \mathcal{N}(0,I), t} \left[ \| \epsilon - \epsilon_\theta(x_t, t, y) \|_2^2 \right]
\end{equation}

where $x_t = \sqrt{\bar{\alpha}_t} \, x + \sqrt{1 - \bar{\alpha}_t} \, \epsilon$ is the noised image at timestep $t$, and $\bar{\alpha}_t = \prod_{s=1}^{t}(1 - \beta_s)$ is the cumulative product of a linear noise schedule $\{\beta_s\}_{s=1}^{T}$. 

To enable classifier-free guidance (CFG) \cite{hoClassifierFreeDiffusionGuidance2022}, the class conditioning is randomly dropped during training with probability $p_{\text{drop}}$, replacing $y$ with a learned null token $\emptyset$, such that the model learns both conditional and unconditional noise prediction jointly.
To obtain the counterfactual $\hat{x}^*$ for input $x$ steered toward the normal class $y_n$, we follow a two-stage procedure. First, the input is partially noised to a timestep $t_{\text{start}} = \lfloor \rho \cdot T \rfloor$, controlled by a noise strength parameter $\rho \in (0, 1)$:

\begin{equation}
    x_{t_{\text{start}}} = \sqrt{\bar{\alpha}_{t_{\text{start}}}} \, x + \sqrt{1 - \bar{\alpha}_{t_{\text{start}}}} \, \epsilon, \quad \epsilon \sim \mathcal{N}(0, I)
\end{equation}

This preserves the coarse structure of the input while introducing sufficient stochasticity for the model to transition toward the target class. The parameter $\rho$ thus governs the tradeoff between structural similarity to the source and the freedom to deviate toward the target class.

Second, we perform a deterministic reverse diffusion from $t_{\text{start}}$ to $0$ using the DDIM sampler \cite{songDenoisingDiffusionImplicit2020}, with CFG applied at each step to steer the trajectory toward $y_n$:

\begin{equation}
    \hat{\epsilon}_t = \epsilon_\theta(x_t, t, \emptyset) + s_{\text{cfg}} \cdot \bigl[ \epsilon_\theta(x_t, t, y_n) - \epsilon_\theta(x_t, t, \emptyset) \bigr]
\end{equation}

where $s_{\text{cfg}} \geq 1$ is the CFG scale, which controls how strongly the denoising trajectory is steered toward $y_n$. The predicted clean image at each step is then:

\begin{equation}
    \hat{x}_0^{(t)} = \frac{x_t - \sqrt{1 - \bar{\alpha}_t} \cdot \hat{\epsilon}_t}{\sqrt{\bar{\alpha}_t}}
\end{equation}

and the DDIM update yields $x_{t-1}$ from $\hat{x}_0^{(t)}$ and $\hat{\epsilon}_t$ without reintroducing stochasticity. The final counterfactual is the terminal denoised image $\hat{x}^* = \hat{x}_0^{(0)}$. The parameters $\rho$ and $s_{\text{cfg}}$ jointly govern how strongly the output is pushed toward the target class, and can be tuned depending on the degree of class transition required. An overview of the whole process can also be seen in Fig.~ \ref{fig:setup_diff}.
As commonly done, we suppress small, diffuse changes from the diffusion model. We interpolate between input and counterfactual using a soft mask obtained by Gaussian-smoothing $|x - \hat{x}^*|$, clipping it at a fraction $\alpha$ of its maximum (with $\alpha$ = 1.0 on the Chest X-ray and Brain MRI data sets, and $\alpha$ = 0.4 on the manometry data set), and normalizing it to $[0, 1]$. Strongly changed regions are thus kept, while weakly changed ones are pulled back toward the input.

\subsubsection{Optimization details}

As described in section \ref{sec_method_cf_diff}, the counterfactual is found by adding noise to the input image and gradually denoise it towards the target class. Table \ref{tab_optimization_hyperparameters_diff} shows the hyperparameters used to create the counterfactual. The parameters were chosen based on multiple runs and deemed suitable enough for our experiments.

\begin{table}
\small
\caption{Optimization parameters}
\label{tab_optimization_hyperparameters_diff}
\centering
\begin{tabular}{@{\hskip 0.1em}l @{\hskip 0.2em} c @{\hskip 0.8em} c @{\hskip 0.8em} c  @{\hskip 0.2em}}
\toprule
 & \multicolumn{3}{c}{\textbf{Datasets}} \\
\cmidrule(r){2-4} 
\textbf{Parameter} & \textbf{Manometry} & \textbf{Chest X-ray} & \textbf{Brain MRI} \\
\midrule
Steps & 50 & 50 & 50 \\
$s_{\text{cfg}}$ & 15 & 7 & 7 \\
$\rho$ & 0.7 & 0.2 & 0.7 \\
\bottomrule
\end{tabular}
\end{table}

\section{Details of data sets}\label{sec:appendix_dataset}

\textbf{Manometry} High-resolution manometry (HRM) measures esophageal pressure during swallowing via a sensor-equipped catheter, producing pressure maps. It is the gold standard for diagnosing swallowing disorders. The data set, collected at TUM University Hospital Munich and annotated by medical specialists, includes six classes representing different contraction force categories.\footnote{Ethical approval was obtained to use the collected data} Missingness is crucial in this data set, as several classes are defined by absent or reduced pressure.

\textbf{Chest X-ray} The data set\footnote{Available at: 
\scriptsize \url{https://www.kaggle.com/datasets/vbookshelf/pneumothorax-chest-xray-images-and-masks/data}}
comprises chest X-rays for pneumothorax classification and segmentation tasks \cite{siim-acr-pneumothorax-segmentation}. A pneumothorax, describing air in the pleural space, can appear as a dark region due to collapsed lung tissue, making missingness a relevant feature in this data set as well. The data set contains two classes, comprising normal and pathological samples.

\textbf{Brain MRI} The data set\footnote{Available at: 
\scriptsize \url{https://www.kaggle.com/datasets/mateuszbuda/lgg-mri-segmentation/data}} 
comprises brain MRIs for tumor classification and segmentation tasks \cite{budaAssociationGenomicSubtypes2019}. The data set contains two classes, comprising normal and pathological samples, with the tumors mostly visible as bright regions due to contrast agent.

The number of training, test and evaluation samples for each of the three data sets can be seen in Table \ref{tab:datasetdescription}. Train/test splits were either predefined (Chest X-ray, Brain MRI) or randomly selected (Manometry).

\begin{table}
\small
  \centering
  \caption{Details of data sets}
 \begin{tabular}{@{\hskip 0.4em} l @{\hskip 0.5em} c @{\hskip 0.5em} c @{\hskip 0.5em} c @{\hskip 0.4em}}
    \toprule
      & \textbf{Manometry} &  \textbf{Chest X-Ray} & \textbf{Brain MRI}\\
    \midrule
    Training samples & 3805  & 10675  & 3126 \\
    Test samples & 922 & 1372 & 773 \\
    Distinct classes & 6 & 2 & 2 \\
    \hdashline[2pt/2pt]
    \noalign{\vskip 0.3ex}
     Evaluation samples & 80 & 263 & 277 \\
    \bottomrule
    \end{tabular}
  \label{tab:datasetdescription}
\end{table}

\section{Analysis of computational cost}\label{sec:appendix_computational_cost}

While training the VAE and diffusion model introduces an upfront training time cost (where the exact cost is depending on the used hardware and training details), it is amortized over inference. The optimization of the latent code for a counterfactual is lightweight and performed for each input image during inference. We conducted measurements regarding the training times of the diffusion and VAE models and the inference time to compute the counterfactual baselines. The results, all of which were computed on a single NVIDIA L40S GPU, can be seen in Table \ref{tab:computational_time_training} and Table \ref{tab:computational_time_inference}.
These results show that compute time during inference to generate a single counterfactual is very low (in the magnitude of at most $\sim$~2.5 seconds). Furthermore, for the VAE approach we observe that the majority of compute time goes into optimizing within the latent space to find the counterfactual latent representation $z^*$. Sampling multiple perturbations around $z^*$ for the EG~(CF) case is then very efficient with almost no overhead. For the Diffusion model, the EG~(CF) case takes more time, since every counterfactual is computed separately through the denoising process.

While this paper focuses purely on IG and extensions of it, we want to still give a brief section how the computational cost compares to other non-IG methods (e.g. SHAP). While it is true that SHAP and related methods do not require an explicit baseline in the same way as IG, they come with their own significant computational demands. For example, SHAP is based on computing Shapley values by evaluating all possible permutations of feature subsets. This leads to exponential complexity ($O(m \cdot 2^m)$) in the number of features $m$, making it intractable for high-dimensional inputs like images without approximations. Even approximations (such as KernelSHAP or DeepSHAP \cite{lundbergUnifiedApproachInterpreting2017}) still require multiple forward passes per sample, often far exceeding the number of passes $k$ needed for IG (which has complexity $O(k)$). This means that our method, while introducing a one-time cost for training the generative model, maintains a low and consistent computational demand during inference. Once the baseline is generated, IG is computed the same way as with any other traditional baseline using a fixed number of steps (as is common practice).

\begin{table}
\small
\centering
\caption{Training time for the generative models (in minutes)}
\label{tab:computational_time_training}
\centering
\begin{tabular}{@{\hskip 0.4em}l @{\hskip 1.8em} c @{\hskip 1.4em} c @{\hskip 1.4em} c  @{\hskip 0.4em}}
\toprule
& \multicolumn{3}{c}{Data sets} \\
\cmidrule(r){2-4}
 & \textbf{Manometry} & \textbf{Chest X-ray} & \textbf{Brain MRI} \\
\midrule
Diffusion & 683.40 & 2436.73 & 747.63 \\
VAE & 33.24 & 65.34 & 28.21 \\
\bottomrule
\end{tabular}
\end{table}

\begin{table}
\scriptsize
\centering
\caption{Computational time during inference to create counterfactual in seconds (mean $\pm$ std)}
\label{tab:computational_time_inference}
\centering
\begin{tabular}{@{\hskip 0.2em}l @{\hskip 1.7em} l @{\hskip 1.2em} l @{\hskip 1.2em} l@{\hskip 0.2em}}
\toprule
& \multicolumn{3}{c}{Data sets} \\
\cmidrule(r){2-4}
\textbf{Approach} & \textbf{Manometry} & \textbf{Chest X-ray} & \textbf{Brain MRI} \\
\midrule
\textsf{CF\textsubscript{Dif.}} (single counterfactual) & 0.74 $\pm$ 0.09 & 0.09 $\pm$ 0.01 & 0.72 $\pm$ 0.00 \\
\textsf{EG (CF)\textsubscript{Dif.}} (50 counterfactuals) & 31.96 $\pm$ 2.24 & 3.76 $\pm$ 0.00 & 31.36 $\pm$ 0.03 \\
\textsf{CF\textsubscript{VAE}} (single counterfactual) & 2.44 $\pm$ 1.06 & 0.06 $\pm$ 0.04 & 2.15 $\pm$ 1.12 \\
\textsf{EG (CF)\textsubscript{VAE}} (50 counterfactuals) & 2.47 $\pm$ 1.06 & 0.07 $\pm$ 0.04 & 2.18 $\pm$ 1.12 \\
\bottomrule
\end{tabular}
\end{table}

\section{Model descriptions}
In this section we provide the specifications of the models that are used in our work (i.e. the Diffusion model, VAE and classifier), as well as the training details.

\subsection{Classifier}\label{sec:appendix_classifier}

\paragraph{Architecture}
Table \ref{appendix_classifier_architecture} shows the abstracted architecture of the classifier that was used in our experiments. We use a swin\_v2\_t architecture with pretrained weights (ImageNet) as the classifier backbone. One classifier model was trained on one of the three data sets each.

\begin{table}
\scriptsize
\caption{Abstracted architecture of the classifier.}
\centering
\begin{tabular}{c @{\hskip 0.1in} c @{\hskip 0in} c}
\toprule
\textbf{Layer} & \textbf{Type} & \textbf{Description} \\
\midrule
Input & Preform Block & Conv2d layer that prepares input channels \\[3ex]

Backbone & SwinTransformerV2 & \\[3ex]

Output & Classification Heads & \makecell[c]{Three attention pooling heads; \\ one shallow and one deeper \\ linear classification head with \\ specified output dimensions;} \\
\bottomrule
\end{tabular}
\label{appendix_classifier_architecture}
\end{table}

\paragraph{Training details}

\begin{figure*}
\centering
\includegraphics[width=\textwidth]{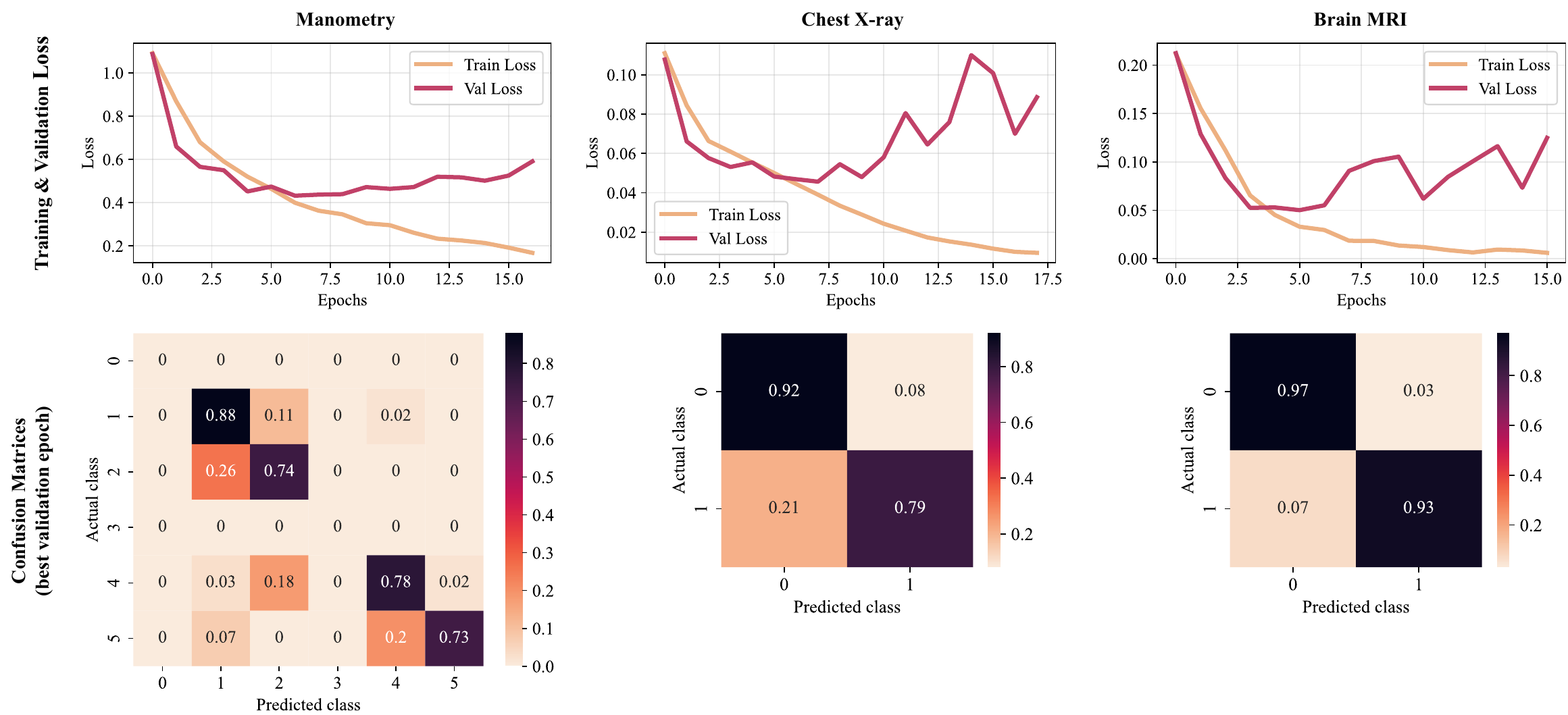}
\caption{Loss on training and validation sets during training of the classifier (first row).  Confusion matrices on the validation sets using of the model from the epoch with the best results on the validation set (second row).}
\label{fig_appendix_classifier}
\end{figure*}

The model is trained using early stopping and we use the model from the epoch with the best accuracy on the validation set. The model is trained to output two separate classifications. The first is a binary classification, indicating if any disease is present or not, and the second one is a classification for the exact class of disease. (Since in the Chest X-ray data set and the Brain MRI data set only one disease is available, both heads are returning the same information). The classifiers are trained using a cross entropy loss function, weighted by the class counts to account for class imbalance. The optimizer used is Adam, with a cosine learning rate schedule.
Fig.~\ref{fig_appendix_classifier} shows the loss curves during training of the classifier, as well as the confusion matrix of the model from the epoch with the best results on the validation set.
Additional hyperparameters are specified in Table \ref{tab_classifier_hyperparameters}. All hyperparameters were chosen based on few experimental runs.

\begin{table}
\footnotesize
\caption{Training parameters of classifier.}
\label{tab_classifier_hyperparameters}
\centering
\begin{tabular}{c @{\hskip 0.1in} c @{\hskip 0.05in} c @{\hskip 0.05in} c}
\toprule
 & \multicolumn{3}{c}{\textbf{Datasets}}                   \\
\cmidrule(r){2-4} 
\textbf{Parameter} & \textbf{Manometry} & \textbf{Chest Xray} & \textbf{Brain MRI} \\
\midrule
Learning rate & 3e-5 & 3e-5 & 3e-5 \\
Weight decay & 1e-4 & 1e-4 & 1e-4 \\
Min. nr. of epochs & 5 & 5 & 5 \\
Max. nr. of epochs & 50 & 50 & 50 \\
Cosine scheduler warmup & 4 & 4 & 4 \\
\bottomrule
\end{tabular}
\end{table}

\subsection{Diffusion model}\label{sec:appendix_diffusion}

\paragraph{UNet architecture} The UNet-based denoising network $\epsilon_\theta$ follows a standard encoder-decoder structure with skip connections. The encoder progressively downsamples the input through a series of ResBlocks with strided convolutions, while the decoder upsamples back to the original resolution via transposed convolutions, with each decoder stage receiving the corresponding encoder feature map via channel-wise concatenation. A self-attention block is inserted at the bottleneck to capture global dependencies. At each stage, the timestep and class label are injected into the ResBlocks via adaptive group normalisation (AdaGN), which modulates the feature maps through a learned scale and shift derived from the combined timestep and label embedding. The architecture is summarised in Table \ref{appendix_unet_architecture}.

\paragraph{Training details}

\begin{figure*}
\centering
\includegraphics[width=\linewidth]{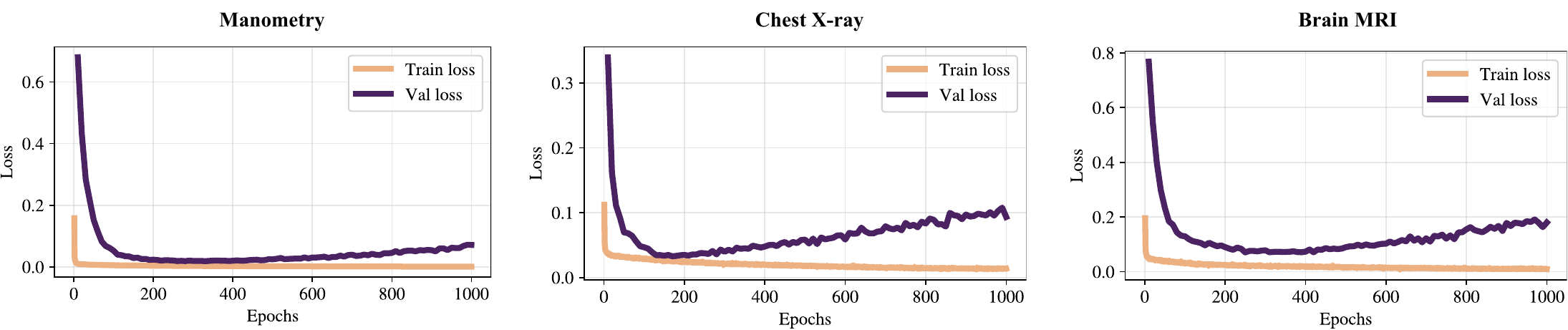}
\caption{Loss on training and validation sets during training of the Diffusion model on the three data sets.}
\label{fig_appendix_diffusion}
\end{figure*}

The diffusion models were trained using the AdamW optimizer and the hyperparameters specified in Table \ref{tab_diffusion_hyperparameters}. The hyperparameters were chosen based on a simple grid search. Fig.~\ref{fig_appendix_diffusion} shows the loss on the training and validation sets during training of the diffusion model on the three data sets.

\begin{table}
\scriptsize
\caption{Training parameters of the Diffusion model}
\label{tab_diffusion_hyperparameters}
\centering
\begin{tabular}{c @{\hskip 0.1in} c @{\hskip 0.05in} c @{\hskip 0.05in} c}
\toprule
 & \multicolumn{3}{c}{\textbf{Datasets}}                   \\
\cmidrule(r){2-4} 
\textbf{Parameter} & \textbf{Manometry} & \textbf{Chest Xray} & \textbf{Brain MRI} \\
\midrule
Input shape & 128$\times$128 & 128$\times$128 & 128$\times$128 \\
Learning rate & 0.0001 & 0.0001 & 0.0001 \\
Batch size & 64 & 64 & 64 \\
Nr. of epochs & 1000 & 1000 & 1000 \\
Drop prob. (classifier-free guidance) & 0.1 & 0.1 & 0.1 \\
Timesteps & 100 & 100 & 100 \\
\bottomrule
\end{tabular}
\end{table}

\begin{table}
\scriptsize
\caption{Training parameters of VAE.}
\label{tab_vae_hyperparameters}
\centering
\begin{tabular}{c @{\hskip 0.1in} c @{\hskip 0.05in} c @{\hskip 0.05in} c}
\toprule
 & \multicolumn{3}{c}{\textbf{Datasets}}                   \\
\cmidrule(r){2-4} 
\textbf{Parameter} & \textbf{Manometry} & \textbf{Chest Xray} & \textbf{Brain MRI} \\
\midrule
Input shape & 512$\times$512 & 512$\times$512 & 512$\times$512 \\
Learning rate & 0.001 & 0.001 & 0.001 \\
Batch size & 256 & 256 & 256 \\
Reconstruction loss & L2 & L1 & L2 \\
$\beta$ (weight of KL-Loss) & 2 & 2 & 4 \\
Nr. of epochs & 200 & 200 & 200 \\
Latent dimension & 512 & 512 & 512 \\
\bottomrule
\end{tabular}
\end{table}

\subsection{VAE}\label{sec:appendix_vae}

\paragraph{Encoder architecture} The encoder architecture of the VAE used for the Manometry and Brain MRI data sets is depicted in Table \ref{appendix_encoder_architecture}, while the encoder architecture used for the Chest X-ray data set is depicted in Table \ref{appendix_encoder_architecture_spatial}. The main difference is the structure of the latent dimension, which is not flattened but instead kept spatial for the latter. This architecture was chosen for the Chest X-ray data set due to better reconstruction results.

\paragraph{Decoder architecture}  The decoder architecture of the VAE used for the Manometry and Brain MRI data sets is depicted in Table \ref{appendix_decoder_architecture}, while the decoder architecture used for the Chest X-ray data set is depicted in Table \ref{appendix_decoder_architecture_spatial}.

\paragraph{Training details}

\begin{figure*}
\centering
\includegraphics[width=\textwidth]{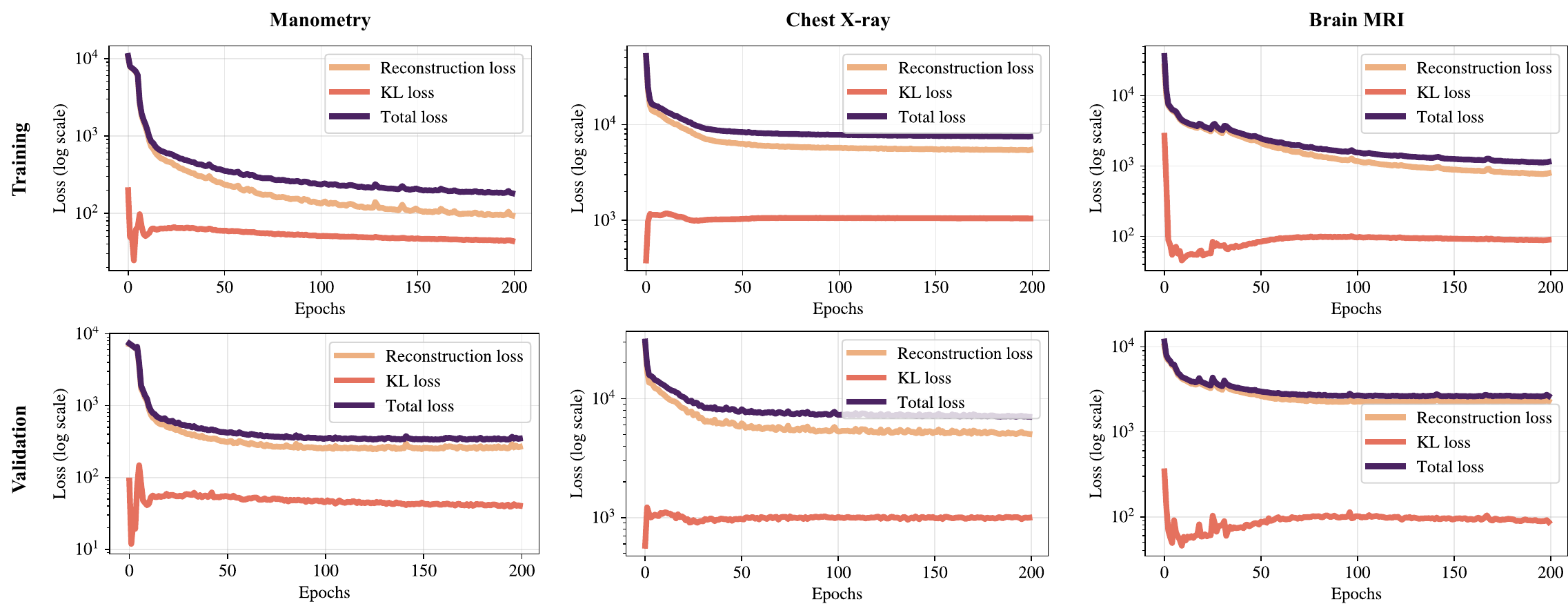}
\caption{Loss on training and validation sets during training of the VAE model on the three data sets.}
\label{fig_appendix_vae}
\end{figure*}

The VAEs were trained using the Adam optimizer and the hyperparameters specified in Table \ref{tab_vae_hyperparameters}. The loss function is combination of the reconstruction loss and a $\beta-$weighted KL-Loss. The hyperparameters were chosen based on a simple grid search. Fig.~\ref{fig_appendix_vae} shows the loss on the training and validation sets during training of the VAE on the three data sets.

\begin{table*}
\footnotesize
\caption{High-level UNet architecture of the diffusion denoising network $\epsilon_\theta$. Each ResBlock consists of two $3\times3$ convolutions with GroupNorm and SiLU activations, conditioned via AdaGN on the combined timestep and label embedding.}
\centering
\begin{tabular}{l @{\hskip 0.55in} l}
\toprule
\textbf{Stage} & \textbf{Type} \\
\midrule
Input conv & $3\times3$ conv, $1 \rightarrow 128$ channels \\
\addlinespace
\multicolumn{2}{l}{\textit{Conditioning}} \\
Timestep & Sinusoidal embedding, 2$\times$ Linear + SiLU, $d_{\text{emb}} = 256$ \\
Label & Learned embedding table(s), Linear + SiLU, $d_{\text{emb}} = 256$ \\
\addlinespace
\multicolumn{2}{l}{\textit{Encoder} (repeated 3 times, channels doubling each level)} \\
ResBlock & $128 \times 2^{l-1} \rightarrow 128 \times 2^{l}$ channels, AdaGN conditioned \\
Downsample & $3\times3$ conv, stride 2 \\
\addlinespace
\multicolumn{2}{l}{\textit{Bottleneck}} \\
ResBlock + Attention + ResBlock & Self-attention between two AdaGN-conditioned ResBlocks \\
\addlinespace
\multicolumn{2}{l}{\textit{Decoder} (repeated 3 times, channels halving each level)} \\
Upsample & $2\times2$ transposed conv, stride 2 \\
ResBlock & $128 \times 2^{l}$ + skip $\rightarrow 128 \times 2^{l-1}$ channels, AdaGN conditioned \\
\addlinespace
\multicolumn{2}{l}{\textit{Output}} \\
GroupNorm + SiLU + $1\times1$ conv & $128 \rightarrow 1$ channel \\
\bottomrule
\end{tabular}
\label{appendix_unet_architecture}
\end{table*}

\begin{table*}
\footnotesize
\caption{Encoder architecture of the convolutional VAE.}
\centering
\begin{tabular}{c @{\hskip 0.35in} c @{\hskip 0.35in} c}
\toprule
\textbf{Layer} & \textbf{Type}  & \textbf{Activation} \\
\midrule
Input & --  & -- \\
Conv1 & $4\times4$ conv, input $\rightarrow$ 32 channels, stride 2, padding 1 & LeakyReLU \\
Conv2 & $4\times4$ conv, 32 $\rightarrow$ 64 channels, stride 2, padding 1 & LeakyReLU \\
Conv3 & $4\times4$ conv, 64 $\rightarrow$ 128 channels, stride 2, padding 1 & LeakyReLU \\
Conv4 & $4\times4$ conv, 128 $\rightarrow$ 256 channels, stride 2, padding 1 & LeakyReLU \\
Conv5 & $4\times4$ conv, 256 $\rightarrow$ 512 channels, stride 2, padding 1 & LeakyReLU \\
Flatten & -- & -- \\
$\text{FC}_{\mu}$ & Fully Connected & -- \\
$\text{FC}_{\sigma}$ & Fully Connected & -- \\
\bottomrule
\end{tabular}
\label{appendix_encoder_architecture}
\end{table*}

\begin{table*}
\footnotesize
\caption{Decoder architecture of the convolutional VAE.}
\centering
\begin{tabular}{c @{\hskip 0.35in} c @{\hskip 0.35in} c}
\toprule
\textbf{Layer} & \textbf{Type}  & \textbf{Activation} \\
\midrule
Input & -- & -- \\
ConvTranspose1 & $4\times4$ conv, 512 $\rightarrow$ 256 channels, stride 2, padding 1 & LeakyReLU \\
ConvTranspose2 & $4\times4$ conv, 256 $\rightarrow$ 128 channels, stride 2, padding 1 & LeakyReLU \\
ConvTranspose3 & $4\times4$ conv, 128 $\rightarrow$ 64 channels, stride 2, padding 1 & LeakyReLU \\
ConvTranspose4 & $4\times4$ conv, 64 $\rightarrow$ 32 channels, stride 2, padding 1 & LeakyReLU \\
ConvTranspose5 & $4\times4$ conv, 32 $\rightarrow$ input channels, stride 2, padding 1 & LeakyReLU \\
Sigmoid & -- & -- \\
\bottomrule
\end{tabular}
\label{appendix_decoder_architecture}
\end{table*}

\begin{table*}
\footnotesize
\caption{Encoder architecture of the convolutional VAE with spatial latent space.}
\centering
\begin{tabular}{c @{\hskip 0.35in} c @{\hskip 0.35in} c}
\toprule
\textbf{Layer} & \textbf{Type} & \textbf{Activation} \\
\midrule
Input & -- & -- \\
Conv1 & $4\times4$ conv, input $\rightarrow$ 32 channels, stride 2, padding 1 & LeakyReLU \\
Conv2 & $4\times4$ conv, 32 $\rightarrow$ 64 channels, stride 2, padding 1 & LeakyReLU \\
Conv3 & $4\times4$ conv, 64 $\rightarrow$ 128 channels, stride 2, padding 1 & LeakyReLU \\
Conv4 & $4\times4$ conv, 128 $\rightarrow$ 256 channels, stride 2, padding 1 & LeakyReLU \\
Conv5 & $4\times4$ conv, 256 $\rightarrow$ 512 channels, stride 2, padding 1 & LeakyReLU \\
$\text{Conv }{\mu}$ & $1\times1$ conv, 512 $\rightarrow$ $C{z}$ channels & -- \\
$\text{Conv }{\log\sigma^2}$ & $1\times1$ conv, 512 $\rightarrow$ $C{z}$ channels & -- \\
\bottomrule
\end{tabular}
\label{appendix_encoder_architecture_spatial}
\end{table*}

\begin{table*}
\footnotesize
\caption{Decoder architecture of the convolutional VAE with spatial latent space.}
\centering
\begin{tabular}{c @{\hskip 0.35in} c @{\hskip 0.35in} c}
\toprule
\textbf{Layer} & \textbf{Type} & \textbf{Activation} \\
\midrule
Input & Spatial latent tensor $z \in \mathbb{R}^{C_z \times H_z \times W_z}$ & -- \\
Conv1 & $3\times3$ conv, $C_z \rightarrow 512$ channels, stride 1, padding 1 & LeakyReLU \\
ConvTranspose1 & $4\times4$ conv, 512 $\rightarrow$ 256 channels, stride 2, padding 1 & LeakyReLU \\
ConvTranspose2 & $4\times4$ conv, 256 $\rightarrow$ 128 channels, stride 2, padding 1 & LeakyReLU \\
ConvTranspose3 & $4\times4$ conv, 128 $\rightarrow$ 64 channels, stride 2, padding 1 & LeakyReLU \\
ConvTranspose4 & $4\times4$ conv, 64 $\rightarrow$ 32 channels, stride 2, padding 1 & LeakyReLU \\
ConvTranspose5 & $4\times4$ conv, 32 $\rightarrow$ input channels, stride 2, padding 1 & Sigmoid \\
\bottomrule
\end{tabular}
\label{appendix_decoder_architecture_spatial}
\end{table*}

\end{document}